\documentclass[journal]{IEEEtran}
\usepackage[numbers]{natbib}
\usepackage[pdftex]{graphicx}
\DeclareGraphicsExtensions{.pdf}
\usepackage[cmex10]{amsmath}
\usepackage{amssymb}
\usepackage{algorithmic}
\usepackage[linesnumbered,ruled]{algorithm2e}
\usepackage{enumerate}
\usepackage{array}
\usepackage{color}
\usepackage{multirow}
\usepackage{adjustbox}
\usepackage[table,xcdraw]{xcolor}
\usepackage{textcomp}
\usepackage{booktabs,makecell}
\usepackage[justification=centering]{caption}
\usepackage{slashbox}
\usepackage{pifont}
\usepackage{lipsum}
\usepackage{xcolor}
\usepackage{xparse}
\NewDocumentCommand{\myrule}{O{1pt} O{3pt} O{black}}{%
  \par\nobreak 
  \kern\the\prevdepth 
  \kern#2 
  {\color{#3}\hrule height #1 width\hsize} 
  \kern#2 
  \nointerlineskip 
}
\usepackage{url}
\usepackage{graphicx, caption}
\usepackage[labelformat=simple]{subcaption}

\usepackage{mathtools,amssymb,lipsum, nccmath}

\usepackage{cuted}
\usepackage[markup=underlined]{changes}
\definechangesauthor[color=purple]{MJ}

\begin{document}
\title{A Spatio-temporal Track Association Algorithm Based on Marine Vessel Automatic Identification System Data}

\author{Imtiaz Ahmed$^{1}$,  Mikyoung Jun$^{2}$ and Yu Ding$^{3}$
\thanks{$^{1}$Imtiaz Ahmed with the Department of Industrial \& Management Systems Engineering,
        West Virginia University,
        Morgantown, WV.
        {Email: \tt\small imtiaz.ahmed@mail.wvu.edu}}%
\thanks{$^{2}$Mikyoung Jun with the Department of Mathematics,
        	University of Houston,
	       Houston, TX.
	{Email: \tt\small mjun@central.uh.edu }}%
\thanks{$^{3}$Yu Ding with the Department of Industrial \& Systems Engineering,
	Texas A\&M University,
	College Station, TX.
	{Email: \tt\small yuding@tamu.edu}}%
}
\maketitle

\begin{abstract}

Tracking multiple moving objects in real-time in a dynamic threat environment is an important element in national security and surveillance system. It helps pinpoint and distinguish potential candidates posing threats from other normal objects and monitor the anomalous trajectories until intervention. To locate the anomalous pattern of movements, one needs to have an accurate data association algorithm that can associate the sequential observations of locations and motion with the underlying moving objects, and therefore, build the trajectories of the objects as the objects are moving.  In this work, we develop a spatio-temporal approach for tracking maritime vessels as the vessel's location and motion observations are collected by an Automatic Identification System. The proposed approach is developed as an effort to address a data association challenge in which the number of vessels as well as the vessel identification are purposely withheld and time gaps are created in the datasets to mimic the real-life operational complexities under a threat environment. Three training datasets and five test sets are provided in the challenge and a set of quantitative performance metrics is devised by the data challenge organizer for evaluating and comparing resulting methods developed by participants. When our proposed track association algorithm is applied to the five test sets, the algorithm scores a very competitive performance.\\
\end{abstract}


\section{Introduction}
\label{sec:intro}
\IEEEPARstart{I}n modern days we are blessed with technologically advanced, sophisticated position tracking systems that can transmit the positions and movement of multiple objects in real time. However, we still lack effective spatio-temporal algorithms that can process this large amount of information gathered over time, helping us build, group and predict trajectories, and in some cases, associate tracking points to their true tracks in the absence of object identification information. The idea of associating or assigning unlabeled moving objects to their true tracks is known as \emph{track association}, extremely important for identifying threats in the form of anomalies. Tracking moving objects using both space and time information to detect anomalous trajectory patterns has far reaching safety implications for maritime security. Towards this end, the National Geospatial-intelligence Agency (NGA), in collaboration with the National Science Foundation (NSF)'s Algorithms for Threat Detection (ATD) program, recently launched data association challenges \cite{website2020}, and design a set of challenge problems using the maritime vessel tracking data, collected by the vessels' Automatic Identification System (AIS). The research reported in this paper is an algorithm developed to address the ATD data association challenge problems. We want to stress that the focus of this work is on associating the AIS measurements/detections to true vessels while the measurements have already been collected by the AIS. We will explain the datasets in more detail in Section~\ref{sec:data}.

AIS is an automated maritime vessel locating, tracking, and monitoring system. Each ship is equipped with a transponder and uses a common very-high-frequency radio channel to communicate with other vessels, the shore-based stations, and some global positioning satellites. AIS was developed under the guidance of the International Maritime Organization (IMO) and is required on most ocean-going commercial ships. According to IMO 2002 convention, international voyaging ships, with 300 or more gross tonnage, and all passenger ships regardless of size require installation of AIS. The primary purpose of AIS used to be helping avoid unwanted collisions among vessels and thereby enhance maritime safety. Nowadays, however, AIS is increasingly used as a monitoring system to track vessel movements at the open sea or near a port \cite{ford:2018}, because AIS can locate another vessel more effectively under reduced or zero visibility than the radar detection system, and thus enable watch guards to prepare in advance \cite{ou:2008}. For this reason, AIS also helps with maritime traffic management and the maintaining of coastal security.

AIS data transmitted by each vessel are in the form of time-sequenced nodes. Each node represents a row or a data point in the AIS data file and contains the timestamps (timestamps marking the signal receiving time of the AIS receiver), coordinates (latitude and longitude), speed (in knots), and direction of a vessel (in degrees). Time sequenced nodes enable receiving groups of messages from an input source, and preserve the order in which the messages in each group arrived. AIS data also include the vessel's maritime mobile service identity (MMSI) number,  unique for each ship. Despite the technological advancement, in the reality of AIS data transmission, collection, and retrieval, the MMSI number could be missing or messed up. When MMSI is absent, it presents a particular challenge for track association. Simply put, the question is that when the MMSI number is removed or missing, could one still associate these time-sequenced nodes correctly in an AIS dataset to recreate the trajectories of the vessels?

Apart from the absence of MMSI, there exist other issues further complicating track association. One such issue is the presence of time gaps in the vessel reporting system. A vessel can randomly stop sending signals for various reasons, ranging from sudden equipment failure to deliberate hiding of the current trajectory. It would be much difficult to associate the nodes to the right track after a long absence of its signal, because during this period, a ship can alter its direction completely, increase the speed dramatically, or even stop altogether. Another difficulty in track association arises when the vessels are near a port. In a port, lots of vessels are parked nearby, making them hard to be differentiated. They also maneuver frequently for parking or for making space for other vessels. Therefore, tracking individual vessels near ports needs special care.

The academic contributions of this work can be summarized as follows:

\begin{enumerate}
\item[\ding{111}] We propose a spatio-temporal track association algorithm that can associate vessel measurements collected from the AIS to their true tracks in real-time. 
\item[\ding{111}] We propose a novel dissimilarity score metric that compares the predicted location of existing vessels with the current AIS measurement in hand and decides between opening a new track versus assigning the node to an existing track in a complete online fashion. We develop our location prediction algorithm based on Vincentry's direct geodesic framework \cite{vincenty:1975}.
\item[\ding{111}] To address the sensitivity of the online, real-time association to some complex tracking phenomena (e.g., abrupt direction change, parking maneuver near the parking port, absence of AIS signal for a significant period of time etc.), we propose a post-hoc track merging step to prevent the creation of unnecessary new tracks. These thresholds are empirically learned and tuned using the training datasets provided by the challenge organizers.
\item[\ding{111}] We also propose a spatially varying track generation and merging policy. It divides the underlying surveillance area into three spatial regions and accordingly approves new track generation or suggests possible merging.
\end{enumerate}

Our proposed approach has the following strengths. First, it is online in nature, meaning that it associates each time stamped node with a vessel on the fly. Second, it makes use of the spatial information to increase its effectiveness for tracking vessels in the open sea and near a port. Third, it can track vessels when there are time gaps in the AIS signals. Fourth, it embodies a provision for offline correction of some difficult-to-separate tracks. Our proposed method is applied to the five data sets provided by the data association challenge organizers. The performance metrics, as also designed by the challenge organizers, are calculated for the proposed method. There were more than ten teams participating in the ATD data challenge. According to the information made available to us, our algorithm performs the second best as compared with those teams. But we are not authorized to disclose the full list of performance statistics of other teams.  In Section~\ref{sec:comparison}, we therefore present the performance of our proposed method together with that of a sample baseline algorithm provided by the challenger organizers. We also present the tracking and association performance of two popular multi-object tracking approaches and compare their performances with our proposed approach.

The rest of the paper unfolds as follows. Section~\ref{sec:data} describes the data format, variables and characteristics of the training and test data sets. The performance metrics designed will be explained later in Section~\ref{sec:perf}. Section~\ref{sec:literature} highlights some of the existing research areas relevant to the track association problem based on the AIS or trajectory data. In Section~\ref{sec:algorithm}, we present the main idea of our track association algorithm. Section~\ref{sec:comparison} explains the performance evaluation criteria and presents a comparison study of different methods under the same challenge. Finally, we summarise the paper in Section~\ref{sec:conclusion}.

\section{Data Description} \label{sec:data}

In this section, we describe the AIS data sets \cite{website2020} used in the ATD data association challenge. The data comes from the historical database of the Nationwide Automatic Identification System supplied by the United States Coast Guard. The data covers the area around Norfolk, Virginia. Six variables are provided for track association, namely, VID (vessel ID), timestamp (in hh:mm:ss format, where hh represent hours, mm represent minutes, and ss represent seconds), latitude (in degrees), longitude (in degrees), speed (in tenths of knots), and course of direction (in tenths of degrees). As speed and direction are expressed in tenths of their respective units, one needs to divide the raw data by 10 to get the actual value. The VID is referred to as the MMSI number in the preceding section.

\begin{figure*}[tb]
	\captionsetup[subfigure]{aboveskip=0pt}
	\centering
	\begin{subfigure}{0.49\textwidth}
		\centering
		\caption{Graphical depiction of the AIS data. Each letter and color code represent an unique vessel. Direction and width of the arrows represent vessel course and speed respectively.}
		\includegraphics[height=2in]{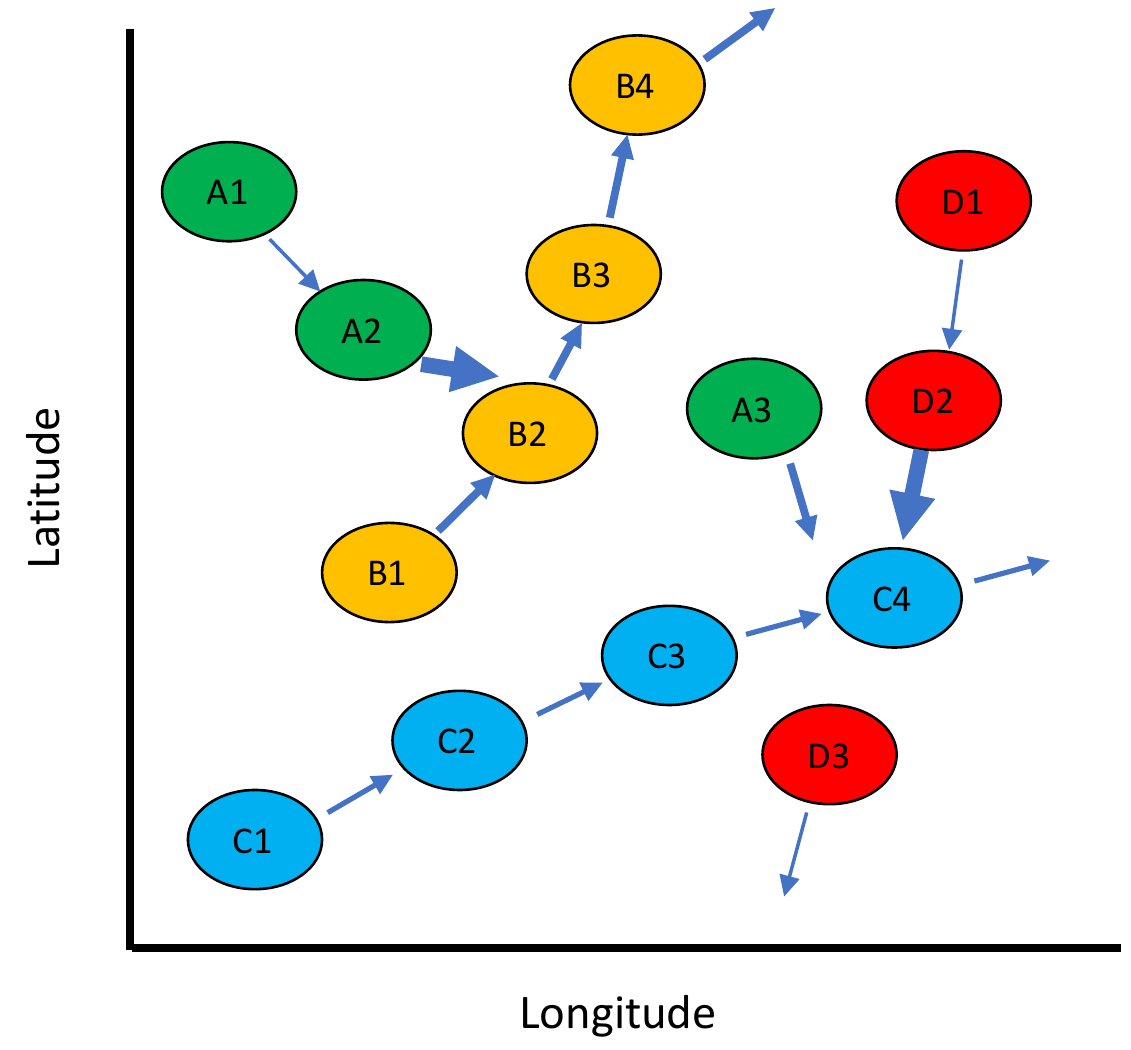}
        \label{fig:AIS1}
	\end{subfigure}
	\begin{subfigure}{0.49\textwidth}
		\centering
		\caption{Tracks consisting of nodes grouped by vessels.}
		\includegraphics[height=2in]{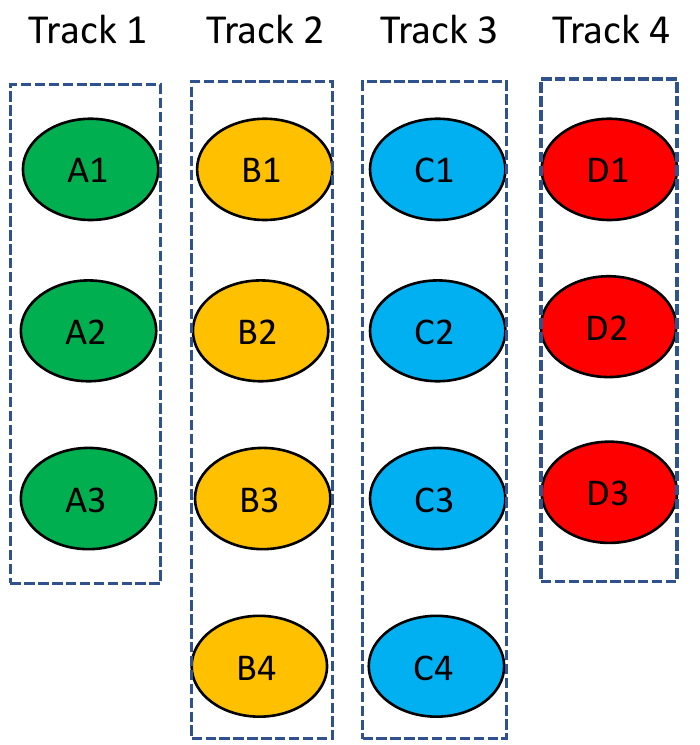}
        \label{fig:AIS2}
	\end{subfigure}
	\begin{subfigure}{0.49\textwidth}
		\centering
		\caption{The AIS data when vessel ID is removed. The numbers represent the objects IDs generated sequentially with respect to their timestamps.}
		\includegraphics[height=2in]{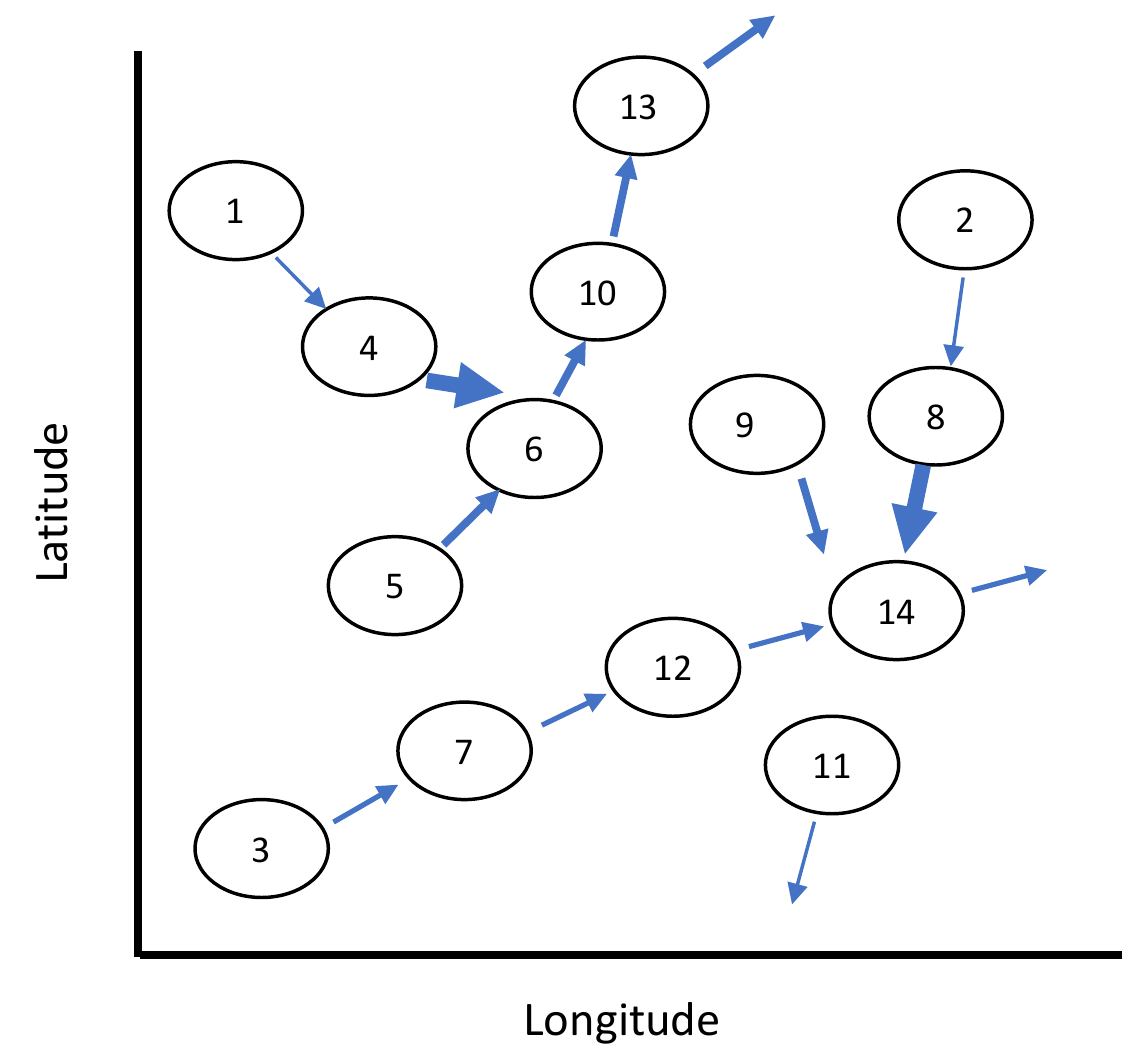}
        \label{fig:AIS3}
	\end{subfigure}
  	\begin{subfigure}{0.49\textwidth}
		\centering
		\caption{Track association in absence of vessel ID.}
		\includegraphics[height=2in]{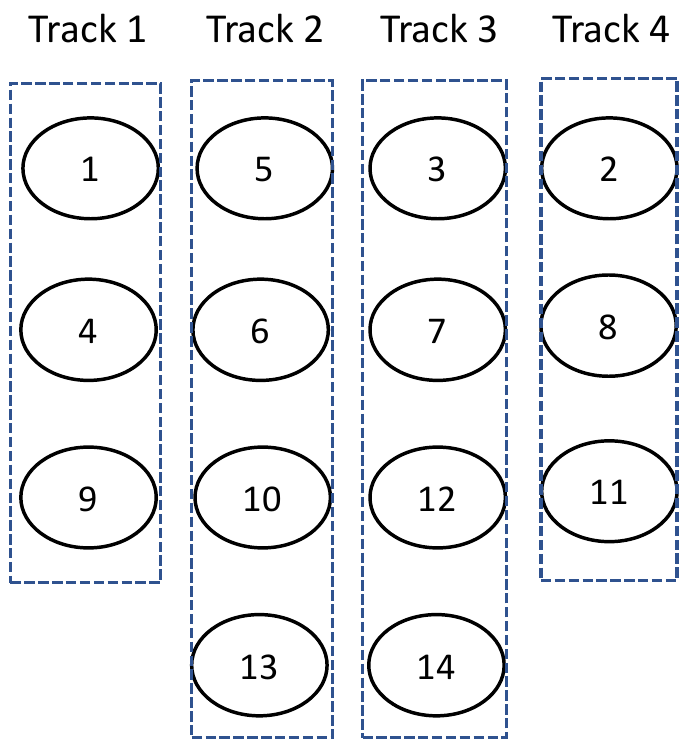}
        \label{fig:AIS4}
	\end{subfigure}
	\caption{Illustration of the track association problem.}
	\label{fig:Total}
\end{figure*}

When the AIS information is received, a monitoring officer can pinpoint it to a single, unique vessel using the VID, if the VID is present and valid. Fig.~\ref{fig:AIS1} and~\ref{fig:AIS2} show an illustration under such circumstance. In Fig.~\ref{fig:AIS1}, each letter (A, B, C or D) represents a unique vessel and the number (1, 2, 3, etc.) that identify the time stamped nodes. Each of these nodes has the associated speed, moving direction, and location in terms of longitude and latitude. Fig.~\ref{fig:AIS2} presents another view, where each of the time stamped nodes is classified into a track associated with a specific vessel. The organization in Fig.~\ref{fig:AIS2} clearly shows the number of unique vessels and how the time stamped information is arranged along each vessel's moving trajectory.

The AIS data received by ships and coastal stations are then transmitted to regional or national data centers. When multiple receivers are connected into the communication and command network, a number of issues could happen, including data intermittency, data redundancy received by multiple receivers, errors in timestamps assigned by varying receivers, and vessels erroneously sharing the message identifier \cite{pallotta:2013}. When the VID information (letter and color code) is missing from the time stamped nodes (see Fig.~\ref{fig:AIS3}), the nodes are then numbered according to the time they are generated. The numbering is sequential in time, with ``1'' meaning the earliest reported node and ``14'' meaning the latest reported node in this example. For example, both A1 and A2 represent the trajectory points of vessel A but A1 appears or reported earlier than A2. Then the desire is to develop an algorithm that can analyze the node information without VID, as in Fig.~\ref{fig:AIS3}, and hopefully group the nodes as in Fig.~\ref{fig:AIS4}, which, if done so, can reveal the same vessel and trajectory information as Fig.~\ref{fig:AIS2} does.

All data files are provided in the comma-separated-values (CSV) format. There are three training datasets, for which the VID associated with each timestamp is given, so that the participants can evaluate the accuracy of their algorithms and then fine tune them. To test the performance of the algorithms generated by the participants, five test datasets are provided by the challenge organizers. The only difference between the training and test datasets is the absence of VID in the test sets. The VID information is intentionally removed so that the competing algorithms can be tested for their association accuracy.


For the purpose of evaluation, competition participants are asked to generate a separate column/variable in the reporting CSV file, known as TRACK\_ID which provides the numbers generated sequentially for each timestamp but unique to a vessel. For example, if we assign the TRACK\_ID `1' to the first record/timestamp and it happens to represent vessel `100008', then each timestamp of the same vessel but appearing later should also be given the TRACK\_ID of `1', if an algorithm works successfully. For each test set, participants are supposed to provide a CSV file with this one extra column of TRACK\_ID.

We summarize the characteristics of the training and test datasets in Table~\ref{tbl:summary}. The training set is referred to as dataset 0 and is further divided into three datasets, i.e., 0-1, 0-2, 0-3, each of which consists of four hours worth of data collected from three separate days. In the training sets, VIDs are known and available to the participants so that they can tune their algorithm parameters to set at optimal values.

Test sets, i.e., datasets 1 to 5, are organized in the increasing order of difficulty. Although the AIS data collection area is somewhat fixed for all five test sets, the track association problem becomes more complicated with the increasing number of vessels and more observations over the five test sets. To add another layer of intricacy, data gap is intentionally incorporated in test sets 4 and 5 by the challenge organizers. This is to mimic the missing data situation often present in real-life datasets. The presence of data gap implies that a selected chunk of the data (e.g., a 20-minute period or several smaller amounts of time for one or more tracks) are removed. All tracks present immediately prior to a gap are still present immediately after the gap, i.e., the track continuity is maintained across each data gap.  For the test sets provided, tracks are expected to keep the same IDs on either side of the gap, i.e., no swaps, breaks or new tracks immediately before and after the gaps. Although there are no track terminations or new initiations immediately across a given gap, such termination and initiation are possible at other times.

The gap characteristics are a bit different between test sets 4 and 5. In test set 4, gaps are created for ALL tracks at two fixed time stamps (not disclosed to the participants), which means that after these particular timestamps, for a certain amount of time no data has been reported from any of the existing vessels. On the contrary, in test set 5, gaps are created only for a subset of the tracks (we do not know which tracks have time gaps). Unlike in set 4, set 5's gaps are not at the same time. Assumingly, it makes the test set 5 more complex than the test set 4. Furthermore, for the test sets, the participants are given a rough estimate of the number of vessels present, not the exact number, and sometimes, only the lower bound.

We want to stress that the datasets are given to us by the Data Challenge Organizer (NGA and NSF). We do not include or incorporate any feature in the dataset by ourselves. We just applied our algorithm to the test sets and then compute the performance metrics which are designed by the challenge oganizers as well. We can guess that they add different features (e.g., time gap, more vessels, more observations) to the test data sets to make the challenge problems more complex. In other words, the Data Challenge specifies strictly what are the inputs and what are the outputs (performance metrics), and all they want is a newer and more capable algorithm in between.

\begin{table*}[h]
\centering
\fontsize{9}{9}\selectfont
\caption{Characteristics of the training and test sets}
\label{tbl:summary}
\begin{tabular}{|l|l|l|l|l|l|}
\hline
\multirow {2}{*}{Problem set}   & VID included  & AIS data & Number of & Number of & Data gap  \\
      &  & duration & vessels & observations &  \\ \hline
0-1 (training) & Yes          & 4 hours                        & 20 &13,714   & No       \\ \hline
0-2 (training) & Yes          & 4 hours                        & 26 &15,707   & No       \\ \hline
0-3 (training) & Yes          & 4 hours                        & 23 &12,940   & No       \\ \hline
1            & No           & 4 hours                       & 8-10    & 3,323           & No       \\ \hline
2            & No           & 4 hours                        & 8-10  & 8,056             & No       \\ \hline
3            & No           & 4 hours                        & \textgreater{}10 & 16,810   & No       \\ \hline
4            & No           & 4 hours                        & \textgreater{}10 & 20,538   & Yes      \\ \hline
5            & No           & 4 hours                       & \textgreater{}10 & 24,513  & Yes      \\ \hline
\end{tabular}
\end{table*}

The target of the AIS data association challenge algorithm is to develop new spatio-temporal association algorithms that can track multiple vessels in a difficult to track scenario and associate the nodes correctly to their respective tracks. Some of the challenges discussed above are purposely introduced so that the existing approaches become much less effective in the association tasks. To provide a quick summary, we further list the following characteristics, which are considered as the bottleneck of the data association challenge:

\begin{itemize}
\item{Each data point is one real object, not a clutter.}
\item{Vessels can pass each other at a very close distance as some of the data corresponds to their parking maneuver in the port. }
\item{The vessel's tracks can cross each other on multiple occasions making it harder to associate the nodes to their corresponding tracks correctly.}
\item{The number of vessels is unknown. Multiple vessels information can be produced at a single time point. The frequency of the incoming information is not fixed and fluctuates through out the information collection period. }
\item{Time gaps are incorporated in the AIS data which means that some of the timestamps are purposely removed from the AIS data. A vessel's trajectory can undergo significant changes during this time gap period thus making the job even harder for the association algorithms.}
\end{itemize}

\section{LITERATURE REVIEW} \label{sec:literature}

There has been a growing interest over the years to collect and analyze the AIS and marine surveillance data to extract important knowledge from the trajectory pattern.  These research works can be categorized into a few major schools of thought, namely route estimation or prediction, trajectory clustering, anomalous pattern identification and multi-object, multi-sensor tracking.

\subsection*{Route Prediction}

Route prediction refers to the prediction of the trajectory pattern of a vessel over time. It includes the estimation of future position and other motion characteristics \cite{pallotta:2013}. It is one of the widely explored research areas involving the AIS data. Route estimation can be further classified into three classes based on their estimation approaches: the physics-based models (motion models)~\cite{caveney:2007,semerdjiev:2000,khan:2005}, the learning-based models~\cite{joseph:2011,pallotta:2014, yang:2022, ma:2022}, and the hybrid models~\cite{perera:2010,stateczny:2011,dalsnes:2018,tu:2020}. It is important to keep in mind that the movement of maritime surface vessels have some unique characteristics, differentiating them from land, air, or even underwater vehicles. For instance, a water surface vessel cannot abruptly stop or change direction \cite{tu:2017}. It spends more time and covers more space during a motion-changing process. Also, a surface vessel moves in a two-dimensional horizontal plane more like a land vessel but unlike air or underwater vessels, which can also move vertically. For the same reason, the tracking task for marine vessels is also different from typical image based object tracking \cite{muresan:2021} or video tracking \cite{wangs:2021}. Luckily, some of these unique characteristics do help predict the route of maritime surface vessels. 


\subsection*{Trajectory Clustering}
Trajectory clustering is to group similar trajectories into clusters. For any clustering model, the single most important factor is to decide on which similarity measure to be used for the purpose of clustering \cite{yuan:2017}. A trajectory is composed of a series of multi-dimensional spatio-temporal points with irregular sampling interval. Traditional similarity measures somewhat fall short of handling this arbitrary discretization \cite{toohey:2015}. As a result, devising various similarity measures, better fit to the vessel trajectory problems, have been a main focus of research effort over the years \cite{chen:2005,vlachos:2006,zhang:2006}. Some popular choices are the principal component analysis (PCA) plus Euclidean distance, Hausdorff distance, Fr{\'e}chet distance, longest common sub-sequence distance (LCSS), and dynamic time warping (DTW). Different algorithms have been tried on trajectory data, including K-means \cite{de:2012}, balances iterative and clustering using hierarchies (BIRCH) \cite{zhang:1996}, density-based spatial clustering of applications with noise (DBSCAN) \cite{wang:2021}, and statistical information grid (STING) \cite{wang:1997}. These trajectory clustering methods can be classified into five groups: spatial-based clustering \cite{jeung:2010,hung:2015,li:2018,sheng:2018}, time-dependent clustering \cite{nanni:2006,mitsch:2013}, partition and group-based clustering \cite{yuan:2012,panagiotakis:2011}, uncertain trajectory clustering \cite{wang:2015}, and semantic trajectory clustering \cite{fileto:2014}.

\subsection*{Anomalous Pattern Identification}

Identification of anomalous patterns from the vessel trajectory data helps locate suspicious activities in the sea \cite{wolsing:2022}. To detect anomalous patterns, the first step is to establish a normalcy baseline, so that a deviation from the baseline can be signaled for further investigation \cite{fu:2017}. The current algorithms for anomalous trajectory pattern detection rely mostly on finding the most representative trajectory from a pool of trajectories, treating it as the normalcy baseline, and then comparing each trajectory with the baseline using a similarity measure \cite{chen:2013,lei:2016}. Anomalies can be tracked to a specific region of the trajectory and features responsible for them. Understandably, similar to the trajectory clustering approaches described above, a suitable dissimilarity measure plays a crucial role in comparing trajectories and signaling potential deviation.


\subsection*{Multi-object Trackers (MOT)}

For maritime security and surveillance, multi-objects trackers (MOT) have also been useful which fuses information from multiple sensors such as radar and sonar rather than directly utilizing the AIS data. MOTs are capable of estimating the future position of moving objects using the past positions of these objects and doing subsequent association. Several tracking algorithms are proposed along this line. Global nearest neighbor (GNN) \cite{blackman:1999} is the simplest algorithm that assigns measurement to track by minimizing a cost assignment matrix. Joint Probabilistic Data Association (JPDA) \cite{bar:1995} is another widely used tracking algorithm where the tracker uses a soft assignment (track association probability) and therefore multiple objects can contribute to the update of each track. Both GNN and JPDA maintain a single hypothesis about the tracked objects. Multiple Hypotheses Tracking (MHT) \cite{reid:1979} is another popular approach for tracking and association which generates a tree of hypotheses for each object. The likelihood of each track is calculated and the most likely combination of tracks is selected. MOT algorithms generally rely on Kalman filter \cite{bishop:2001} to estimate the vessel's dynamic states from positional and directional measurements.

However, MOTs, though seemed a good candidate for vessel tracking and association, are not ideal for handling our AIS challenge datasets. The first issue is that we have cleaned and clutter free AIS object (vessel) information in hand instead of noisy information from sensor(s) that needs to be processed and fused by the MOTs. Second, we can utilize the spatial information of the underlying surveillance area which is different from these conventional MOTs. Third, we believe that the tracking filters used in MOT algorithms are not capable of handling the complex tracking phenomena present in these challenge datasets (e.g., non-linear trajectory, abrupt direction or velocity change between two updates, parking maneuver near the parking port, absence of AIS signal for a significant period of time etc.). When we apply GNN and JPDA to our challenge datasets, they produce poor track association performances (detail in Section V-C).

\section{The Proposed Track Association Algorithm} \label{sec:algorithm}
In this section, we propose our approach for the track association problem. Our approach consists of two separate stages: (1) Online track association and (2) Post-hoc merging. We want to first elaborate the design of our approach, especially the thought process behind the two stages.
At the high level, we, in both stages utilize the unique movement features of the sea vessels (more or less steady course, no abrupt changes in speed and direction) to create detection conditions, and then use empirically learned thresholds to check these conditions and guide track association.

The first stage works in an online manner. It associates each node with a track as it appears along the process. For each node, our online track association algorithm either chooses to open a new track or assigns the node to an existing track. It is built upon a location prediction framework which enables us to estimate the next node location of an existing track and thereby helps anticipate the current node position for deciding its track association. This prediction fits perfectly into the famous Vincenty's direct geodesic problem framework \cite{vincenty:1975}. For that, we make use of a recent solution \cite{karney:2013} of the geodesic problem to estimate the node location. For the association task, we develop a novel dissimilarity metric that helps us to find the right association between a track and the underlying AIS measurement. The solution works quite well when the AIS data is collected at a higher frequency, i.e., the distance traveled by a vessel before the receipt of its next AIS measurement is small, so having this competent solution adds the first competitive edge to our online association approach.

Furthermore, to make our approach robust to real-life operational complexity, we devise a varying threshold policy which takes into consideration the sensitivity of the prediction formula conditioned on the distance a vessel has recently traveled. We find that this conditional threshold policy injects the second competitive edge to our online association.

Despite the much enhanced capability of our online association approach, it is apparent to us that the online algorithm could still make a number of mistakes, especially in terms of creating unnecessary new tracks, due to its myopic nature.  This observation motivates us to design a second stage, after completing the online stage, for deciding whether some of the tracks should be merged. Particular attention in the second stage is given to the spatially varying pattern of new track generation. We adopt policies, learned from the training data, to treat special cases, e.g., significant time gap between two nodes from the same track, and vessels getting in or out of parking locations.  The inclusion of this location-sensitive post-hoc merging in the second stage is the third reason behind the success of our overall track association algorithm.

\subsection{Online Track Association}\label{sec:ota}
Let us represent the set of true tracks as $I$, the set of associated tracks as $J$, and the set of incoming nodes/timestamps as $K$. Understandably, the size of set $I$ and that of set $J$ may not be equal. Each node/timestamp, $k \in K$, comes with the AIS measurements, namely the time of measurement ($t_{k}$), current position ($p_{k}$, which has two components, the latitude, $\phi_{k}$, and the longitude, $\lambda_{k}$), speed ($v_{k}$), and course of direction ($\theta_{k}$). Note that we convert all speed values from knots to meters per second (m/s).

The online track association stage starts off by opening an associated track ($j=1$) and assigning the very first node ($k=1$) to this track. From this point onward, for each incoming node, $k$, we either assign it to one of the existing tracks in set $J$ or open a new track, i.e., add a new element to $J$. We use the notation, $k^{j}$, to indicate that node $k$ is on an associated track $j$.  Following the notation, the AIS measurements of this node $k$ on track $j$ are denoted by $t_{k^{j}}$, $p_{k^{j}} := (\phi_{k^{j}}, \lambda_{k^{j}})$,  $v_{k^{j}}$, $\theta_{k^{j}}$. When both an associated track and a true track appear, we usually use $j$ to denote the associate track and $i$ to denote the true track.  Because of this, unless otherwise indicated, $k^{i}$ is reserved for node $k$ on a true track $i$.

To carry out this online evaluation, we calculate an individual dissimilarity score, denoted by $s_{jk}$ as in \eqref{eq:DISsimi}, to assess the similarity between an existing associated track $j$ and the current node $k$ in consideration. A low score implies high similarity and therefore indicates a possible association to an existing track. For each existing associated track, $j \in J$, its estimated next node measurements are indexed by $n^{j}$, where $n$ implies \emph{next}. The most recent AIS measurements on track $j$ is one step back from $n$ and thus represented by $n^{j}-1$.

Using these notations, we propose the following dissimilarity score:
\begin{align}
    s_{jk}= c_{dist}(p_{k},{p}_{n^{j}}) + c_{ang}(\theta_{k},\theta_{(n^{j}-1)}).
    \label{eq:DISsimi}
    \end{align}
The two terms, $c_{dist}$ and $c_{ang}$, denote changes in distance and angle, respectively, and they are calculated as
\begin{align}
\label{eq:DIST}
&c_{dist}(p_{k},{p}_{n^{j}})= 2r \times\\& \arcsin\sqrt { \sin^{2} \Bigl(\frac{\phi_k-{\phi}_{n^{j}}}{2}\Bigr) + \cos\phi_{k} \cos{\phi}_{n^{j}} \sin^{2} \Bigl(\frac{\lambda_k-{\lambda}_{n^{j}}}{2}\Bigr)}\nonumber
\end{align}
and
\begin{equation}
\label{eq:Ang}
c_{ang}(\theta_{k},\theta_{(n^{j}-1)})=\frac{180- \big|180-| \theta_{k}-\theta_{(n^{j}-1)} |  \big|}{  |t_{k}-t_{(n^{j}-1)}  |}, \end{equation}
where $r$ denotes the Earth's radius. The $c_{dist}$, in \eqref{eq:DIST}, captures the spatial distance between the predicted location and the position of the current node based on the Haversine distance \cite{goodwin:1910}, whereas the $c_{ang}$, in \eqref{eq:Ang} measures the angular change of direction over time between the associated track's last known direction and the direction of the current node and it is devised so as to account for the circular nature of the angular measurements, i.e., the $0$ and $360$ degrees are the same. The spatial proximity ($c_{dist}$) of the current node and the predicted location of existing vessels helps us to decide whether they belong to the same track or not. On the other hand,  $c_{ang}$ tracks the required change in course per unit time to reach to the current location for each existing vessel and thereby indicates the feasibility of grouping them under the same track. It is normally assumed that a vessel does not abruptly change its direction. Together they provide a complete picture of the similarity between the current node and existing tracks.

\begin{figure}[tb]
	\centering
	\centerline{\includegraphics[width=3in, height=2.5in]{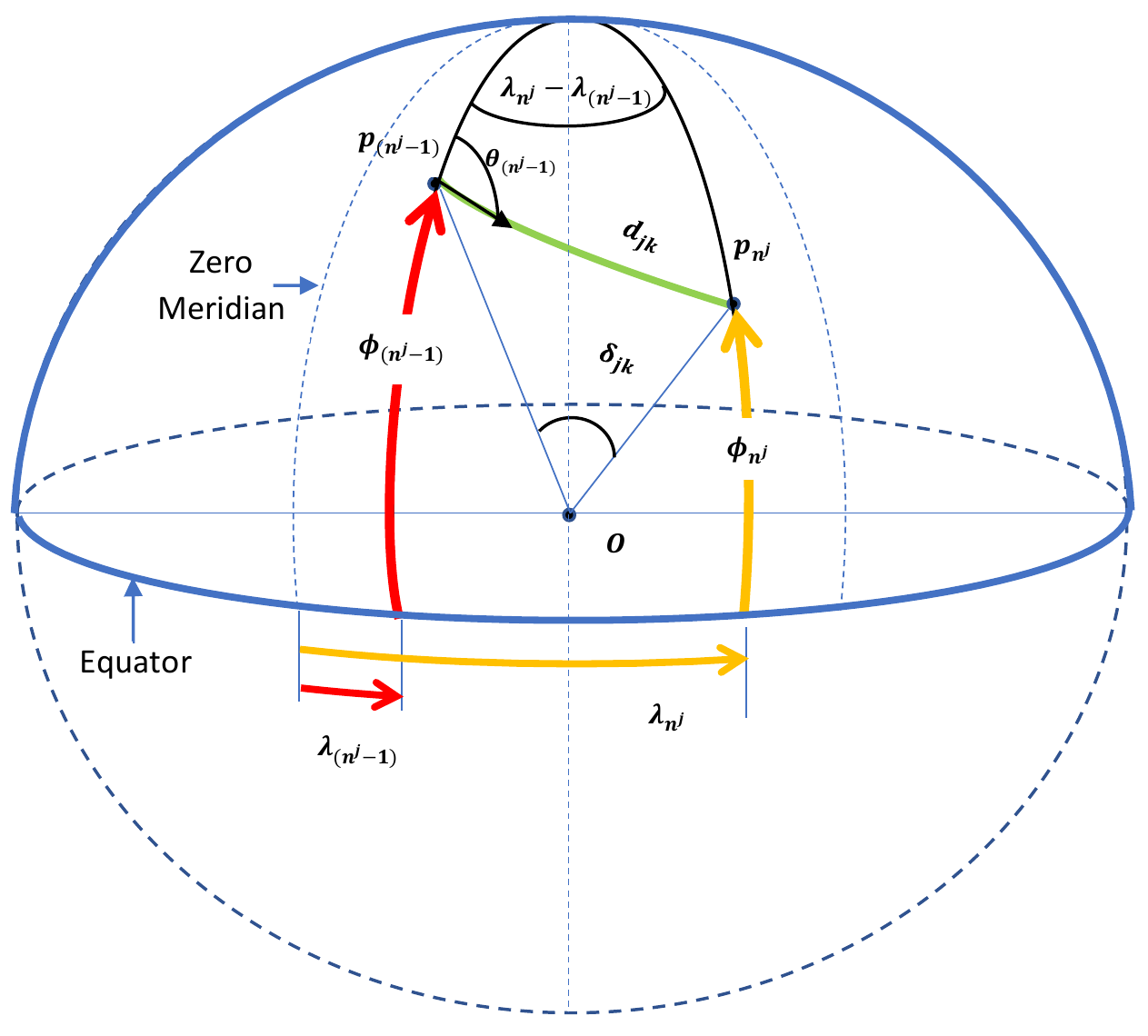}}
\caption{Illustration of predicting the new node location. The green segment represents the geodesic path followed to reach the new location.}
	\label{fig:locc}
\end{figure}

While most of the variables in \eqref{eq:DIST} and \eqref{eq:Ang} are readily available from the AIS measurements, the location coordinates associated with $n^j$, i.e., $p_{n^{j}}:=({\phi}_{n^{j}}, {\lambda}_{n^{j}})$, need to be predicted using its last node's location, $p_{(n^{j}-1)}$, and bearing, $\theta_{(n^{j}-1)}$.  This is done by solving the Vincenty's direct geodesic problem \cite{vincenty:1975}. We skip the details of the derivation but present a graphical illustration in Fig.~\ref{fig:locc}.  The final formula for the two coordinates are:
\begin{align}
\label{eq:loc}
{\phi}_{n^{j}}=&\arcsin(\sin\phi_{(n^{j}-1)}\cos\delta_{jk}+\cos\phi_{(n^{j}-1)} \sin\delta_{jk}\nonumber \\&\times \cos\theta_{(n^{j}-1)}) \end{align}
and
\begin{align}
\label{eq:locs}
{\lambda}_{n^{j}}=\lambda_{(n^{j}-1)}+\arctan (&\sin\theta_{(n^{j}-1)}\sin\delta_{jk}\cos\phi_{(n^{j}-1)},\nonumber \\&\cos\delta_{jk}-\sin\phi_{(n^{j}-1)}\sin\phi_{n^{j}}).
\end{align}
In the above two equations, to measure the angular distance, $\delta_{jk}$, we let
\begin{align}
\label{eq:Delta}
{\delta_{jk}=\displaystyle\frac{d_{jk}}{r},}
\end{align}
where $d_{jk}$ represents the distance traveled along the shortest path on an ellipsoid (the geodesic). We further estimate this $d_{jk}$ through
\begin{equation}
\label{eq:DISTraveled}
d_{jk}= \frac{v_k+v_{(n^{j}-1)}}{2}\times  |t_{k}-t_{(n^{j}-1)}|.
\end{equation}
Note that, the use of the average speed in \eqref{eq:DISTraveled} is an approximation, since the vessel's speed during the time traveled may be quite different (e.g., acceleration, deceleration etc.). Without any additional information to inform us of the vessel's behavior between the two measurements, we find that the use of the average speed provides a good approximation when the distance traveled is small. In this paper, we use a R package called \texttt{geosphere} \cite{hijmans:2017} to generate the values of $p_{n^{j}}$ and calculate the Haversine distance ($c_{dist}$). The package only needs the last known measurements of each vessel, current AIS measurement and the distance traveled to reach to the current location.

Finally, we calculate an overall dissimilarity score for node $k$,
\begin{equation}
\label{eq:Siml}
s_{k}\coloneqq\displaystyle\min_{j \in J} s_{jk},
\end{equation}
which is the minimum of the individual dissimilarity scores $s_{jk}$ as in \eqref{eq:DISsimi} over all $j$'s in the set of $J$. We initially decide to compare $s_{k}$ against a threshold of $\beta$. If $s_{k}>\beta$, a new track would open, and otherwise the node would be assigned to an existing track, say $z \in J$, which returns the lowest dissimilarity score. That is, $z=\displaystyle\operatorname*{argmin}_{j \in J} s_{jk}$ for each $k$.

We find that using the dissimilarity score, $s_k$ as in \eqref{eq:Siml}, there are a couple of complexities that need to be taken are of.  The first issue is a result of having $s_k$ as a summation of both the distance change and the direction change. But recall that the maritime surface vessels are not supposed to change its course of direction abruptly, yet $s_k$, due to its design, is not always sensitive to an angular change alone. For instance, near a port, when vessels are passing each other in close distance with similar speed, then only directional difference can separate these vessels, not the distance and speed. To handle this complexity, we decide to add a threshold, $\alpha$, only on the directional change, $c_{ang}$, in addition to checking $s_k$. When $c_{ang} > \alpha$, then open a new track, regardless of what $s_k$ is; otherwise, chose one of the existing tracks based on $s_k$.

The second issue is how to choose the threshold for $s_k$.  When $s_k$ is very small or very big, the decision of choosing an existing track ($s_k$ small) or opening a new track ($s_k$ large) is reasonably robust, much as expected.  When $s_k$ is in between, we observe that a single threshold on $s_k$, as we originally envisioned, is not robust enough. Another factor, which is the distance traveled, i.e., $d_{zk}$ as in \eqref{eq:DISTraveled}, affects the track association decision. Generally speaking, a large value of $d_{zk}$ usually leads to a relatively less accurate location prediction, and as such, needs a higher $s_{k}$ value for reaching a better decision.

In light of the above discussion, we create a global minimum value of $\beta$, denoted as $\beta_{small}$, and a global maximum value, denoted as $\beta_{large}$, so that if $s_{k}\leq\beta_{small}$, we are going to assign node $k$ to track $z$, i.e., using an existing track, whereas if $s_{k}>\beta_{large}$, we are going to open a new track. When $\beta_{small}<s_{k}\leq\beta_{large}$, we devise another layer of decision, conditioned on $d_{zk}$. To do that, we introduce another threshold of $\mu$. When $\beta_{small}<s_{k}\leq\beta_{large}$, we open a new track if $d_{zk}\leq\mu$, otherwise the node would be assigned to the existing track, $z$.

\begin{figure}[tb]
	\centering
	\centerline{\includegraphics[width=5in, height=6in]{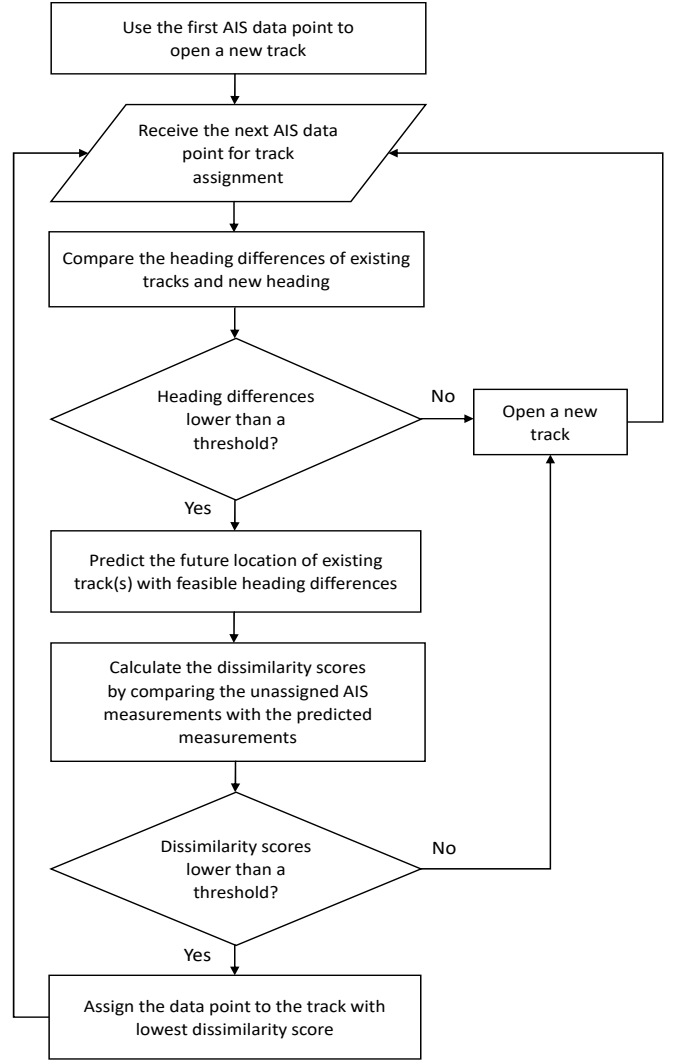}}
\caption{A schematic flow diagram of the online track association process.}
	\label{fig:sch}
\end{figure}

All of these thresholds, i.e., $\beta_{small}$, $\beta_{large}$, $\mu$, and $\alpha$ are determined empirically from the training datasets. For better understanding, a schematic flow diagram of the online track association process is summarized in Fig.~\ref{fig:sch}.

\subsection{ Post-hoc Merging}\label{sec:pm}
The purpose of the post-hoc analysis is to handle the situations that the online clustering does not handle adequately, often due to the need for the online training to keep up with the ongoing process and the limitation of not seeing all the data. Our online clustering method tends to produce more tracks than the number of true tracks, so that our post-hoc analysis does mostly merging than splitting.

To consider the possibility of merging, we create a set of candidate tracks, $A^{(j)}$ for each associated track $j$, where $A^{(j)} \subset J$ and $j \notin A^{(j)}$. Let us represent the start and end node of an associated track $j \in J$ as $s^{j}$ and $e^{j}$, respectively; again the superscript $j$ here signifies the association of $s$ and $e$ with track $j$. Given a track $a \in A^{(j)}$, the end node of track $a$ is thus denoted by $e^{a}$.  For the subsequent analysis, we restrict the tracks in $A^{(j)}$ to be those that stopped before track $j$, i.e., $t_{s^{j}}>t_{e^{a}}$ for $a \in A^{(j)}$.

Our post-hoc merging targets two major problems. The first problem is the wrongful track association due to the presence of time gap. In the presence of such time gap, our location prediction becomes less accurate, and consequently, the online clustering step tends to open a new track whenever a new node appears after the gap, even if it is from an existing track. We consider the following situation as candidates for merging---a candidate track whose last node appeared more than $\tau$ seconds apart from the first node of the new track, yet the two nodes are close than a distance threshold $\gamma$.  The condition is formally stated below:
\begin{equation}
\label{eq:DISTthresh1}
t_{s^{j}}-t_{e^{a}} \geq \tau \,\, \text{and} \,\,  c_{dist}(p_{s^{j}},p_{e^{a}})\leq \gamma, \quad \text{for} \,\, a \in A^{(j)} .
\end{equation}

The second issue to handle is when a vessel changes its direction abruptly, i.e., when $c_{ang}(\theta_{k},\theta_{(n^{j}-1)})$ is large, which leads the online step to open a new track (recall the $c_{ang}>\alpha$ criterion used in the online clustering step). While we institute the angular change condition to open a new track, it could open too many new tracks. But with the angular condition in the online step, the algorithm will miss genuine new tracks. So it is a balance act between over-detection and under-detection.  We find a simpler approach, which is to have the online angular condition as it is but do a post-hoc correction.  The correction is to merge tracks which have their start and end nodes too close to each other, expressed below:
\begin{equation}
\label{eq:DISTthresh2}
c_{dist}(p_{s^{j}}, p_{e^{a}}) \leq \eta.
\end{equation}

Note that whatever issue arises, merging will always take place with a single candidate track, $a$. If there are multiple choices, we will always select the nearest track.

\begin{figure}[tb]
	\centering
	\centerline{\includegraphics[width=3.4in, height=2.4in]{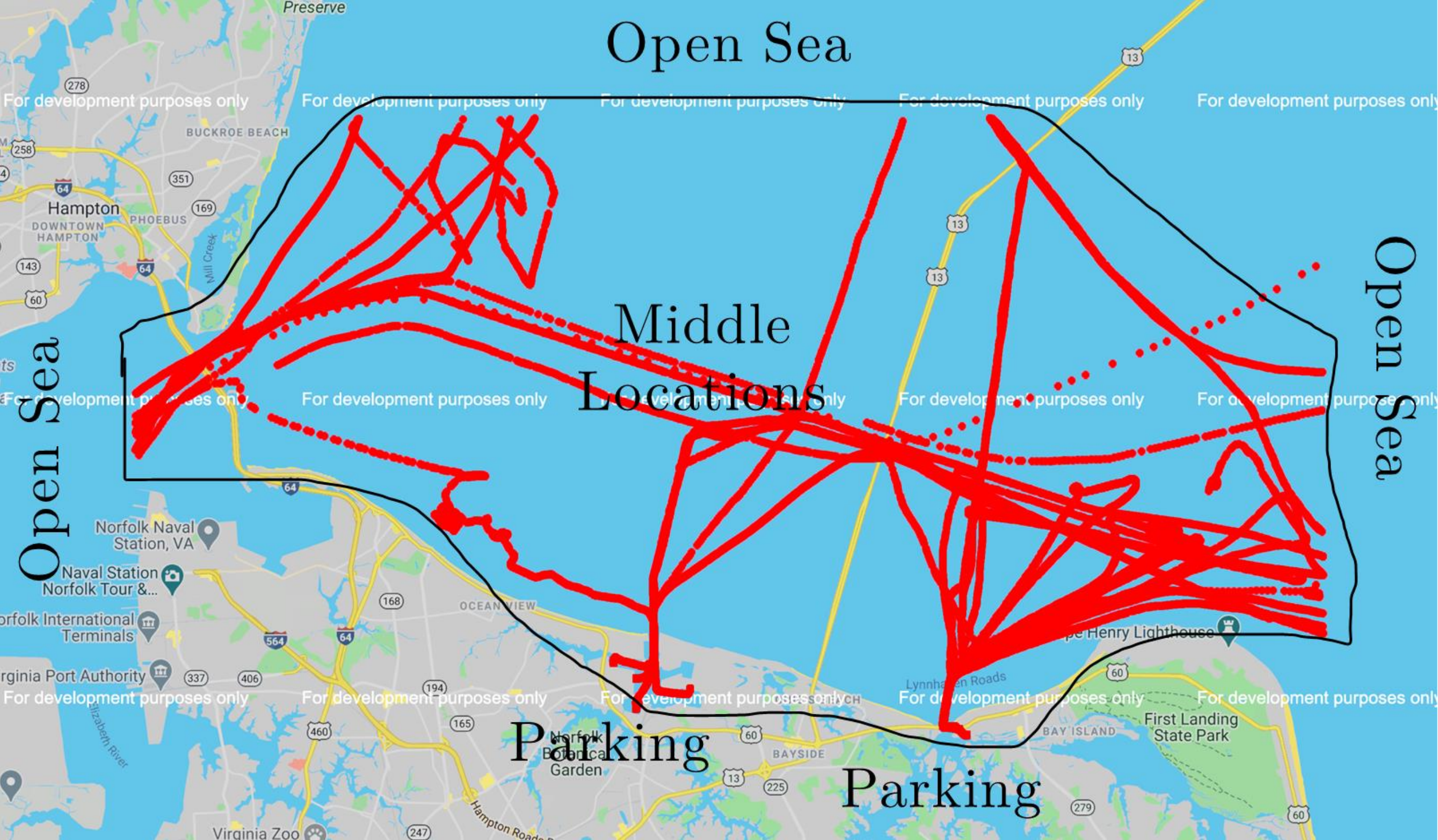}}
\caption{Illustration of boundary, parking and middle locations for track initiation using test set 5. Black line indicates the boundary and the red lines indicates the vessel positions.}
	\label{fig:bn}
\end{figure}

We observe that the likelihood of a post-hoc merging depends heavily on the nature of the locations where a new track emerges. For the datasets provided to the participants, they cover a finite area nearby the Norfolk port.  The vessel locations are truncated, so that there is a clear boundary surrounding the area under monitoring where the AIS data are available; please see Fig.~\ref{fig:bn}. We notice that three types of locations provides important clues for merging likelihood: the boundary locations (in the open sea), the middle locations, and the parking location (which is the boundary locations on land). We devise the following rules of thumb to guide the post-hoc merging based on some common understanding:

\begin{itemize}
\item \textbf{Middle locations}: New tracks cannot appear suddenly out of nowhere. When a new track appears here, it is either the result of a time gap or due to an abrupt direction change in the course of an existing track. We must merge such tracks with the nearby track.
\item \textbf{Boundary locations}: New tracks could emerge any time at the boundary points. As there is no information discerning where the new track comes from, we cannot rule out the possibility of a genuine new track, so such tracks will not be evaluated for merging.
\item {\textbf{Parking locations}}: New tracks can initiate from the parking locations but the historical data show that the initial timestamps of such tracks are always within the first 30 minutes. So we will test a new track for merging only when it emerges after 30 minutes.
\end{itemize}

\begin{figure}[tb]
	\centering
	\centerline{\includegraphics[width=3.5in, height=1.25in]{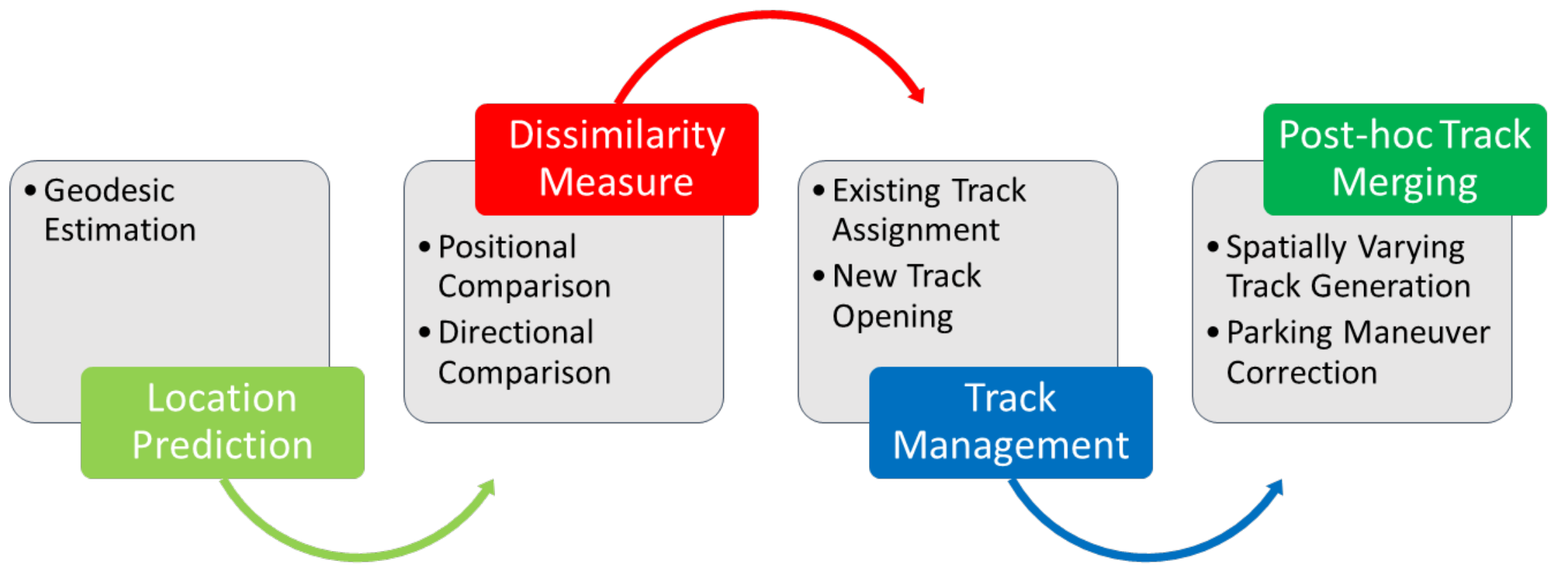}}
\caption{A block diagram to summarize the proposed framework.}
	\label{fig:bl}
\end{figure}

Combining all the elements, the whole algorithm is summarized in Algorithm~\ref{algorithm1}. We also summarized the key components of the entire association process in Fig.~\ref{fig:bl}.

\begin{algorithm}[]
\caption{The proposed track association algorithm}
\label{algorithm1}
\SetKwInOut{Input}{Input}
\SetKwInOut{Output}{Output}
\Input{The set of nodes/timestamps, denoted by $K$ and the initial track set $J = \emptyset$, and six thresholds: $\beta_{{small}}$, $\beta_{{large}}$, $\mu $, $\alpha$, $\tau$, $\gamma$, $\eta$ .}
\Output{Associated track IDs, $j$, for each node}
     \textbf{Stage 1: Online track association}\\
    Open an associated track ($j=1$), and append $j$ to $J$.  Assign the first node ($k=1$) to this track\;
     Starting from $k=2$, \For{each node $k \in K$}{
        \For{each opened associated track $j \in J$}{
    Predict the track's next node location, $p_{n^{j}}$ using \eqref{eq:loc} and \eqref{eq:locs}\;
    Calculate the dissimilarity scores, $s_{jk}$ using \eqref{eq:DISsimi} \;
    Calculate the distance traveled, $d_{jk}$ using \eqref{eq:DISTraveled}\;}
    Calculate the overall dissimilarity score, $s_{k}$ using \eqref{eq:Siml}\;
    Record the track index that return the smallest dissimilarity score, $z=\displaystyle \operatorname*{argmin}_{j \in J} s_{jk}$\;
                \eIf{($\beta_{small}<s_{k}\leq\beta_{large}$ \textbf{and} $d_{zk}$ $\le$ $\mu $) \textbf{or} ($s_{k}$  $>$ $\beta_{large}$) \textbf{or} ($c_{ang}$ $>$ $\alpha$)}{
                    Open a new associated track, append it to $J$ and assign the current node, $k$, as the first element of this new track\;
                   Go to step 3\;
                   }{
                    Assign the current node, $k$, to track $z$\;
                    }}
 \textbf{Stage 2: Post-hoc merging}\\
 \For{each associated track $j \in J$}{
    \eIf{{Starts near the boundary in the open sea} \textbf{or} first member appears within initial 30 minute time window}{
                      Leave the track as is and return to step 19\;
                   }{
                  Find candidate tracks that came before $j$ and save them in $A^{(j)}$, such that $A^{(j)} \subset J$ \textbf{and} $t_{s^{j}}>t_{e^{a}}$ for $a \in A^{(j)}$\;
                  \eIf{($t_{s^{j}}-t_{e^{a}} \geq \tau$ \textbf{and} $c_{dist}(p_{s^{j}},p_{e^{a}})$ $\leq$ $\gamma$) \textbf{or} ($c_{dist}(p_{s^{j}},p_{e^{a}})$ $\leq$ $\eta$)}{
                   Merge track $j$ with track $a$\;
                   Update the set of associated tracks $J$\;
                   Go to step 19\;
                    }{Merging is not feasible and go back to step 19 \;}}}
\end{algorithm}

\section{PERFORMANCE EVALUATION} \label{sec:comparison}

In this section, we evaluate the performance of our proposed track association algorithm on five test datasets. In Section~\ref{sec:perf}, we describe the performance metrics used for the performance evaluation. In Section~\ref{sec:perf1}, we provide the detailed performance of our approach, together with the sample baseline algorithm provided for the data association challenge. In Section~\ref{sec:perf2}, we evaluate the performance of two traditional MOT approaches with respect to our proposed track association algorithm. Finally, in Section~\ref{sec:perf34}, we evaluate the performance of our proposed approach on an open-source AIS dataset.

\subsection{Performance Metrics} \label{sec:perf}
To assess and quantify the performance of a tracking algorithm, the ATD data challenge organizers provided three performance metrics \cite{website2020}, which are described in the sequel.

\subsubsection{Counts of erroneous tracks}\label{sec:basic}

\begin{figure}[tb]
	\centering
	\centerline{\includegraphics[width=3.7in, height=2.8in]{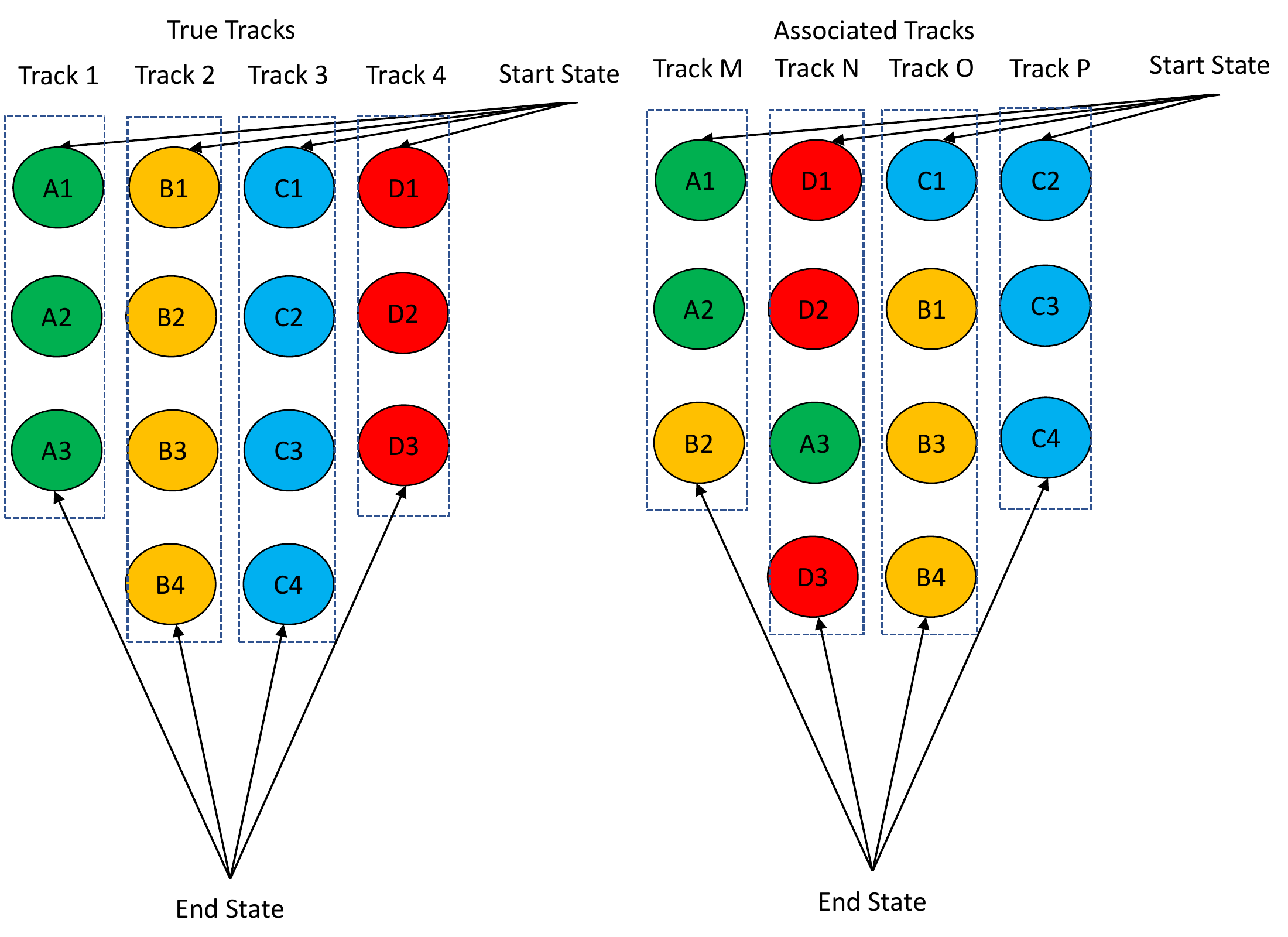}}
\caption{A sample scenario of true and associated tracks, where the start and end states of each track are marked.}
	\label{fig:PM1}
\end{figure}

The purpose of this metric is to measure the quality of track association by counting the number of instances of different errors that could happen during the association. To provide a clear understanding, a simple track association example is presented in Fig.~\ref{fig:PM1}. It has four true tracks (Track 1, Track 2, Track 3, Track 4). Each track consists of multiple nodes, where a pair of consecutive nodes form a segment, e.g., A1 and A2 are the two nodes from Track 1 and A1-A2 is a segment. There are 14 nodes in total, which are distributed into the four tracks. There are also four associated tracks (Track M, Track N, Track O, Track P), resulted from a track association algorithm. As this example is generated to depict different errors, the track association is arbitrarily set and understandably made imperfect. The start and end states of each of these tracks are specifically marked as they play a major role in defining different errors.

We use five error categories to check the quality of a track association algorithm: missed track, extra track, merged track, broken track, and swapped track. Other than the aspect of swapped tracks, we use only the start and end states of the associated tracks to measure these errors. They are defined below:

\begin{itemize}
\item Missed track: A true track's start state that is not captured by any of the associated track's start states. In the sample example, node B1 corresponds to a missed track.
\item Extra track: An associated track's start state that does not match with any of the true track's start states. In the sample example, node C2 corresponds to an extra track.
\item Merged track: A true track's end state that is not captured by any of the associated track's end states. In the sample example, node A3 corresponds to a merged track.
\item Broken track: An associated track's end state that does not match with any of the true track's end states. In the sample example, node B2 corresponds to a broken track.
\end{itemize}

As one only needs the start and end states, these four error categories are also applicable when the number of true and associated tracks are not the same. We do not use the internal nodes in counting these four errors to keep the evaluation process as simple as possible.


To count the number of swapped tracks we have to use the information of track segments. A track swap occurs when a segment on a true track is not on any of the associated tracks. From Fig.~\ref{fig:PM1}, ten true track segments can be identified: A1-A2, A2-A3, B1-B2, B2-B3, B3-B4, C1-C2, C2-C3, C3-C4, D1-D2, and D2-D3. Similarly, ten associated track segments can be defined: A1-A2, A2-B2, D1-D2, D2-A3, A3-D3, C1-B1, B1-B3, B3-B4, C2-C3, and C3-C4. We can see that there are five true track segments (A2-A3, B1-B2, B2-B3, C1-C2, and D2-D3) which are not on any of the associated tracks. So, the number of track swaps in this case is five. The lower bound for all these errors is zero, which represents a perfect association.
Note that using the five errors, as described above, does not present a clear indication of track matching or association. Track association will be done using the completeness score, to be described in Section~\ref{sec:basic}.

\subsubsection{Continuity score}\label{sec:basic}
A continuity score is conceptually the mirror of track swapping. It is defined as the ratio of the distance of all the correct associated track segments to the distance of all the true track segments:
\begin{align}
\label{eq:CS}
&\text{Continuity score}=\frac{\displaystyle\sum_{j\in J} \sum_{k^{j}=2}^{n_{(k^{j})}}g(p_{(k^{j})},p_{(k^{j}-1)})}{\displaystyle\sum_{i\in I} \sum_{k^{i}=2}^{n_{(k^{i})}}c_{dist}(p_{(k^{i})},p_{(k^{i}-1)})}
\end{align}
with
\begin{align}
    g&(p_{(k^{j})},p_{(k^{j}-1)})\nonumber\\&=\begin{cases}
c_{dist}(p_{(k^{j})},p_{(k^{j}-1)}) & \text{ if }[{k^{j}},{(k^{j}-1)}] \in T \nonumber\\
0 & \text{ if } [{k^{j}},{(k^{j}-1)}] \not\in T \nonumber
\end{cases}. \nonumber \\
\end{align}
Here, $k^{i}$ and $k^{j}$ denote running indices for the nodes of true track $i$ and associated track $j$, respectively. Furthermore, $n_{(k^{i})}$ and $n_{(k^{j})}$ represent numbers of total nodes available under true track $i$ and associated track $j$, respectively. The term $T$ represents the set of true track segments. We use the Haversine distance measure, $c_{dist}$, for each true segment as defined in \eqref{eq:DIST}. The continuity score is normalized between 0 and 1, where 1 means perfect. Similar to the preceding five error counts, using the continuity score does not lead to track matching or association, either.

\subsubsection{Completeness score}\label{sec:basic}

Completeness of a true track with respect to an associated track is measured by taking the ratio of the number of nodes that are present in both the concerned true and associated tracks over the number of nodes that are in the true track. If there are multiple such associated tracks available for a true track, then the completeness score of that true track is measured by taking the maximum of all the completeness values, that is,
\begin{equation}
\label{eq:CPS}
\text{Completeness score of true track $i$}=\max_{ j\in J }\frac{b_{ij}}{n_{(k^{i})}}.
\end{equation}
Here, $b_{ij}$ represents the number of nodes on true track $i$ that are also present in associated track $j$. The completeness measure also helps one decide track matching or association, if one uses the associated track attaining the maximum completeness value as the representation of the true track. When track association is done, all the following scenarios are possible: a true track matched to multiple associated tracks, an associated track matched to multiple true tracks, and an associated track not mapped to any true tracks.

\subsection{Comparative Performance Evaluation} \label{sec:perf1}


To apply our algorithm on the test sets, at first, we tune the parameters adopted in our algorithm using the three training datasets for which we know the true VIDs. To summarize, the parameters used in the algorithms are:  $\mu=20$ meters, $\beta_{small}=40$, $\beta_{large}=550$,  $\alpha=25$ degrees/second, $\tau=300$ seconds, $\gamma=3,000$ meters and $\eta=20$ meters. We use the same settings for all five test sets. After we ran our algorithm on the test sets, the challenge organizers provided us the true track information for the test sets \cite{website2020}, so that we can compute the performance metrics described in the preceding section.  We did not change any aspect of our algorithm once we received the true track information of the test sets.

At the end of the data association challenge, the challenge organizers provided us the values of the performance metrics generated by more than ten participating teams, so that we can compare our performance to the best teams.  We are not given the identities of these participants nor the specifics of the approaches used by them. Through this comparison, we know that our method performs overall the second best.  However, we are not authorized to disclose the performance statistics of other teams. Readers who wish to get this information will need to reach out directly to the NSF ATD Program and make a request.

In this section, in addition to present the performance of our proposed method, we include the performance of a sample algorithm \cite{website2020}, which was made available by the challenge organizers to all participants at the beginning of the competition.  We mark our performance values in bold font  whenever it comes out as the best (or tied for the best) in a particular category relative to all participating teams, including that of the sample algorithm.  This allows us to demonstrate the competitiveness of our method without breaching the confidentiality constraint.

Table~\ref{tbl:1} summarizes the count of erroneous tracks representing the number of missing, extra, merged, broken and swapped tracks, produced by our algorithm in five test sets. Inside each parenthesis, it is the corresponding count of erroneous tracks obtained by using the sample baseline algorithm. Here, having a lower count indicates a better performance and zero is the lowest (best) possible value for these counts.

Our algorithm achieves the best performance in four out of five test sets in terms of missed tracks and five out of five cases in terms of extra tracks.  On these two error categories, our method outperforms other participating approaches and ranked the first. As the counts of both missed and extra tracks depend on correctly detecting the opening node of a track, we can say that our approach is the most successful in capturing the start of the tracks. In terms of the counts of merged and broken tracks, which, on the other hand, depends on the capability of marking the end node of a track, our approach again produces superior outcomes and becomes the best method in three test sets and remains very close to the best in the other two sets. In the case of swapped tracks, our approach is the best in two test sets and remains competitive in other three sets (we can confirm our method is in fact ranked the second). In all cases, our method outperforms the sample algorithm comprehensively and by a large margin.

\begin{table*}[h]
\centering
\fontsize{9}{9}\selectfont
\caption{Counts of erroneous tracks produced by our algorithm. Values in the parenthesis are the corresponding count produced by the sample algorithm. Bold indicates the best performance among all participants.}
\label{tbl:1}
\begin{tabular}{|l|l|l|l|l|l|}
\hline
Categories     & Test Set-1 & Test Set-2 & Test Set-3 & Test Set-4 & Test Set-5  \\ \hline
Missed Tracks  & \textbf{0} (1)      & \textbf{0} (1)      & \textbf{0} (7)      & 2 (7)      & \textbf{1} (7)       \\ \hline
Extra Tracks   & \textbf{0} (\textbf{0})      & \textbf{0} (\textbf{0})      & \textbf{0} (11)     & \textbf{2} (12)     & \textbf{1} (14)      \\ \hline
Merged Tracks  & \textbf{0} (6)      & \textbf{0} (5)      & 1 (15)     & \textbf{0} (13)     & 1 (15)      \\ \hline
Broken Tracks  & \textbf{0} (5)      & \textbf{0} (4)      & 1 (91)     & \textbf{0} (18)     & 1 (22)      \\ \hline
Swapped Tracks & \textbf{0} (729)    & \textbf{0} (1,296)   & 169 (6,484) & 44 (7,953)  & 1,250 (8,452) \\ \hline
\end{tabular}
\end{table*}

\begin{table*}[h]
\centering
\fontsize{9}{9}\selectfont
\caption{Continuity and completeness score produced by our algorithm. Values in the parenthesis are the corresponding scores produced by the sample algorithm. Bold indicates the best performance among all participants.}
\label{tbl:2}
\begin{tabular}{|l|l|l|l|l|l|}
\hline
Performance Measures               & Test Set-1    & Test Set-2    & Test Set-3    & Test Set-4    & Test Set-5    \\ \hline
Continuity Score                   & \textbf{1.000} (0.839) & \textbf{1.000} (0.935) & 0.945 (0.636) & 0.988 (0.755) & 0.998 (0.699) \\ \hline
Completeness Score (mean values)   & \textbf{1.000} (0.807) & \textbf{1.000} (0.845) & 0.984 (0.604) & 0.883 (0.628) & 0.916 (0.613) \\ \hline
Completeness Score (median values) & \textbf{1.000} (0.843) & \textbf{1.000} (\textbf{1.000})  & \textbf{1.000} (0.546) & \textbf{1.000} (0.575) & \textbf{1.000} (0.648) \\ \hline
\end{tabular}
\end{table*}

In Table~\ref{tbl:2}, we summarize the continuity scores and completeness scores generated by our algorithm and the sample algorithm in five test sets. A higher value of the completeness scores indicates a better performance with `1' being the best. A continuity score of `1' indicates that all the true track segments are perfectly captured by the associated track segments.

We further breakdown the completeness score into the mean completeness scores and the median completeness scores.  Both completeness scores represent the central tendency of the completeness scores across the numerous tracks in a test set.  Because the distribution of the completeness score is not symmetric, the two values do not agree.  Their disagreement suggests a skewness towards the high end of the score, i.e., $\text{completeness}=1$. Moreover, a mean completeness score of `1' indicates that all of the nodes from each true track are perfectly captured by an associated track, and a median completeness score of `1' indicates that more than half of the true tracks are completely captured by the associated tracks. In the category of median completeness scores, our algorithm produces the best result in all five test sets.  In the continuity scores and mean completeness scores, our algorithm produces the best result in two test sets and very competitive results in the other three test sets (again we confirm that ours is ranked the second). Also, we again comprehensively outperform the sample algorithm in all categories across all five test sets and our performance is on top with a comfortable margin.

Finally, in Table~\ref{tbl:9}, we show the number of associated tracks identified by our algorithm and the sample algorithm, as compared to the actual number of tracks. Here, our approach identified the number of tracks correctly for all five test cases.

\begin{table*}[tb]
\centering
\fontsize{9}{9}\selectfont
\caption{The number of associated tracks recovered by our algorithm and the sample algorithm. Our algorithm recovered perfectly all true track numbers which were unknown to us prior to the application of our method.}
\label{tbl:9}
\begin{tabular}{|l|l|l|l|l|l|}
\hline
Algorithms    & Test Set-1 & Test Set-2 & Test Set-3 & Test Set-4 & Test Set-5 \\ \hline
True Number   & 8          & 10          & 25          & 25          & 25          \\ \hline
Our Algorithm & \textbf{8} & \textbf{10} & \textbf{25} & \textbf{25} & \textbf{25} \\ \hline
Sample Algorithm       & 7          & 9           & 29          & 30          & 32          \\ \hline
\end{tabular}
\end{table*}

\begin{figure*}[tb]
	\captionsetup[subfigure]{aboveskip=0pt}
	\centering
	\begin{subfigure}{0.49\textwidth}
		\centering
		\caption{Test set 1.}
		\includegraphics[height=2in]{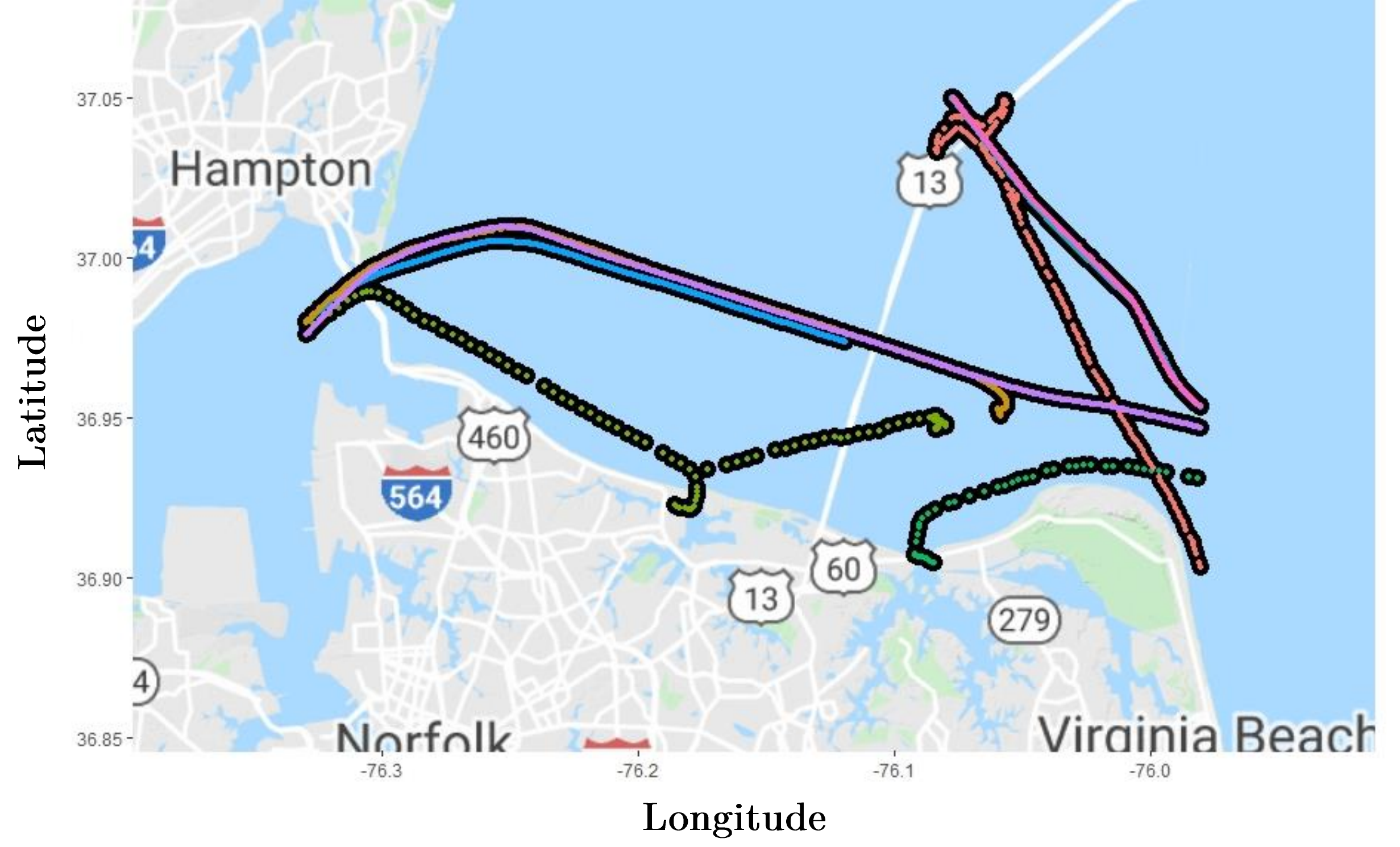}
        \label{fig:AISs1}
	\end{subfigure}
	\begin{subfigure}{0.49\textwidth}
		\centering
		\caption{Test set 2.}
		\includegraphics[height=2in]{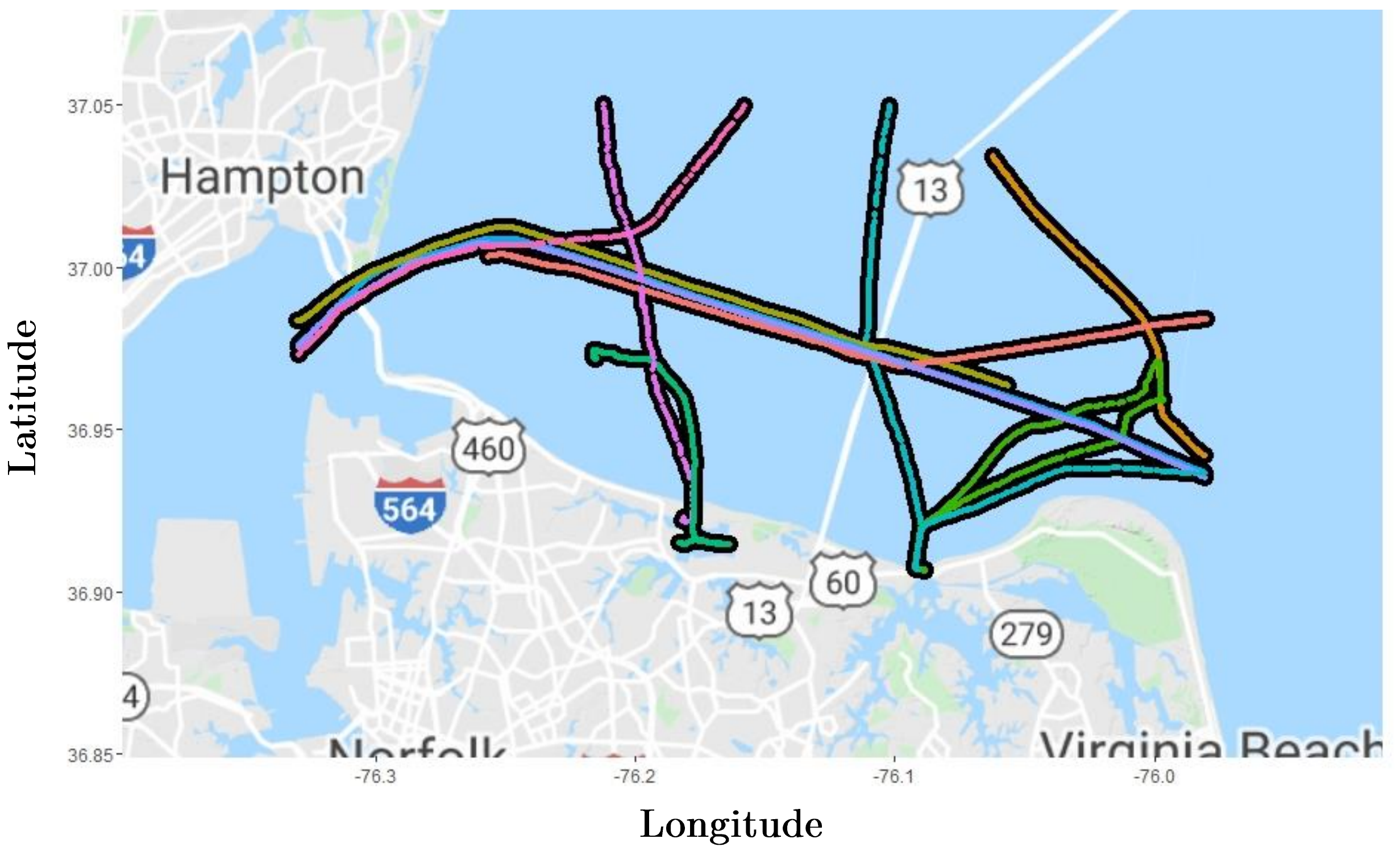}
        \label{fig:AISs2}
	\end{subfigure}
	\begin{subfigure}{0.49\textwidth}
		\centering
		\caption{Test set 3.}
		\includegraphics[height=2in]{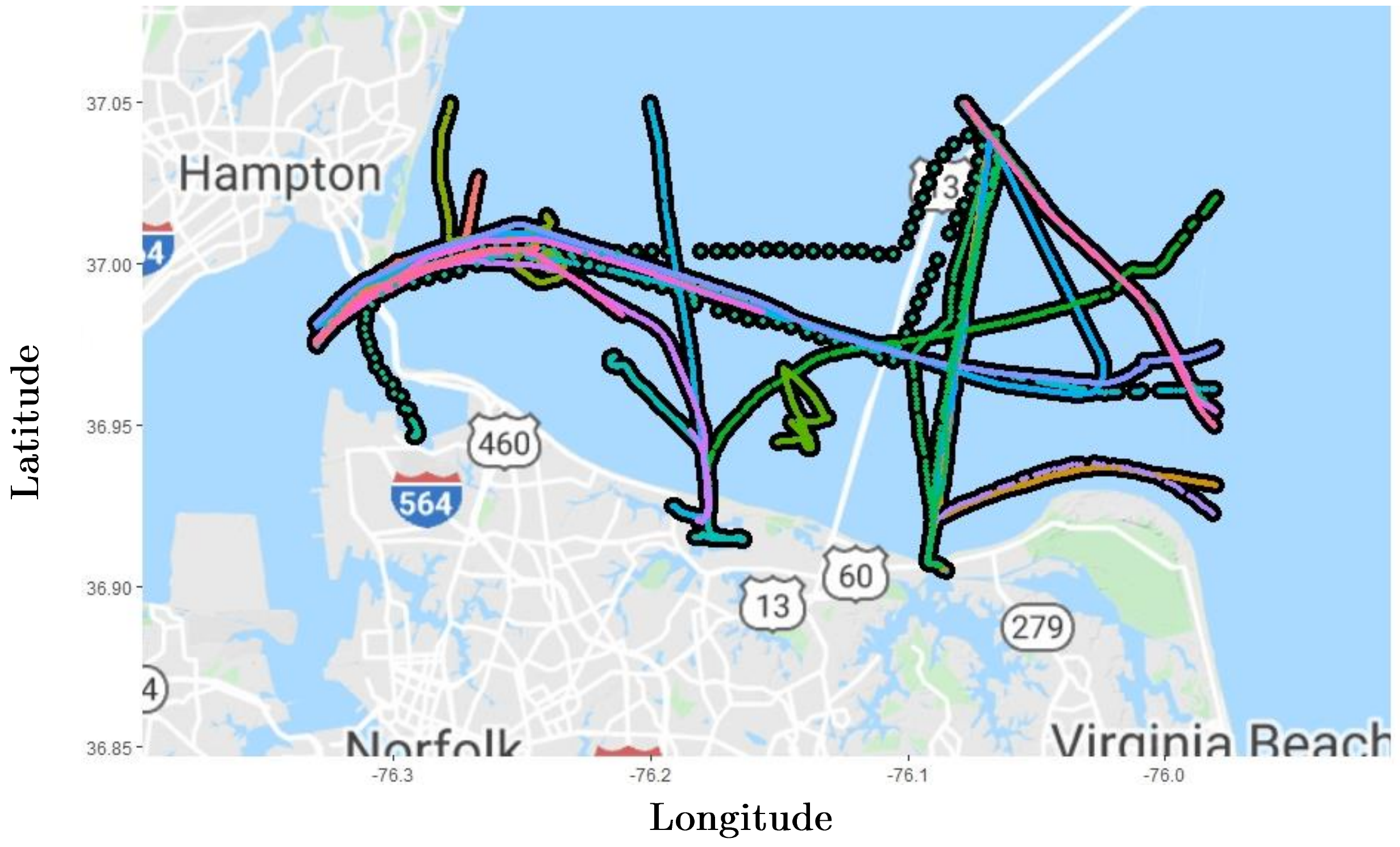}
        \label{fig:AISs3}
	\end{subfigure}
  	\begin{subfigure}{0.49\textwidth}
		\centering
		\caption{Test set 4.}
		\includegraphics[height=2in]{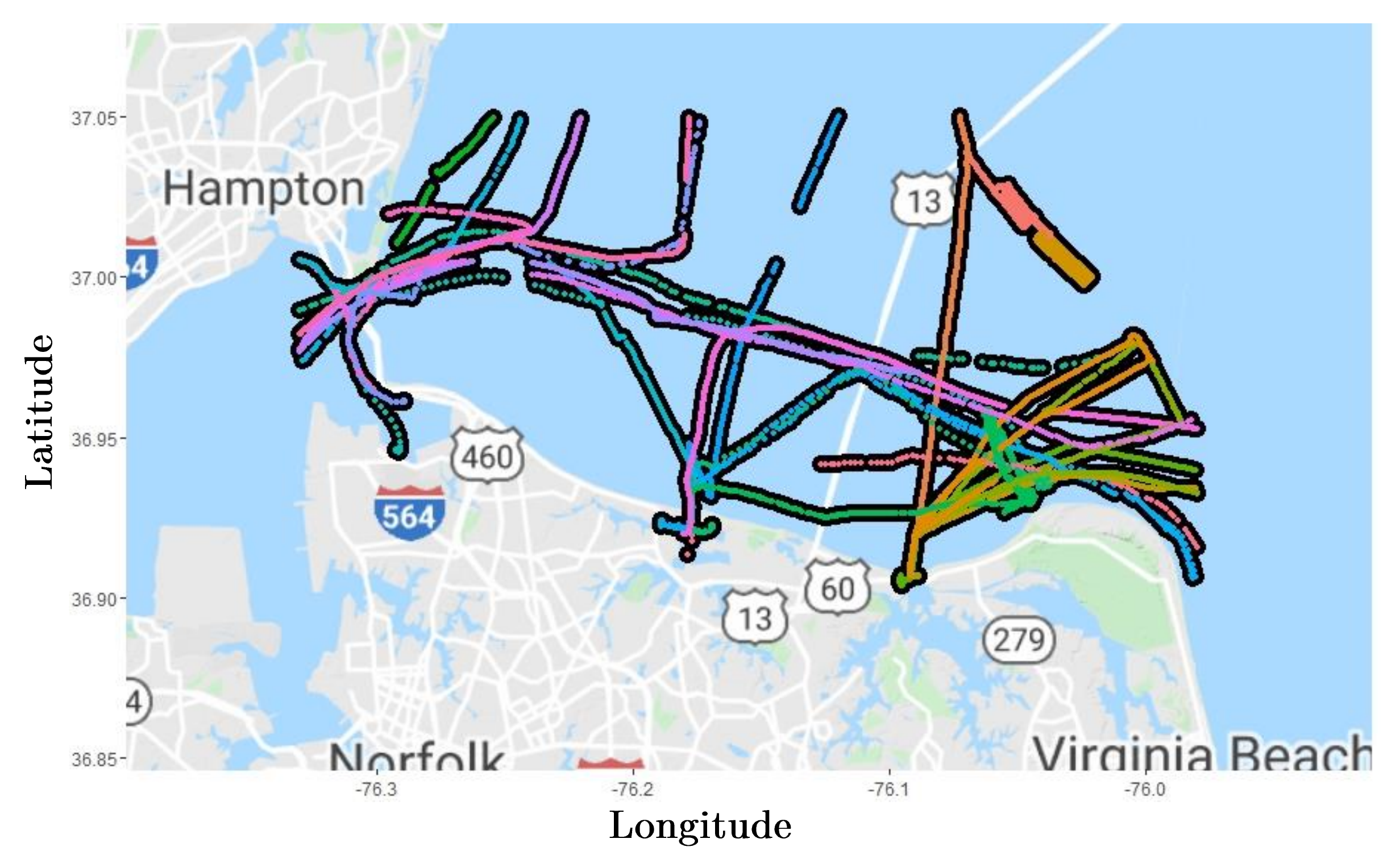}
        \label{fig:AISs4}
	\end{subfigure}
  	\begin{subfigure}{0.49\textwidth}
		\centering
		\caption{Test set 5.}
		\includegraphics[height=2in]{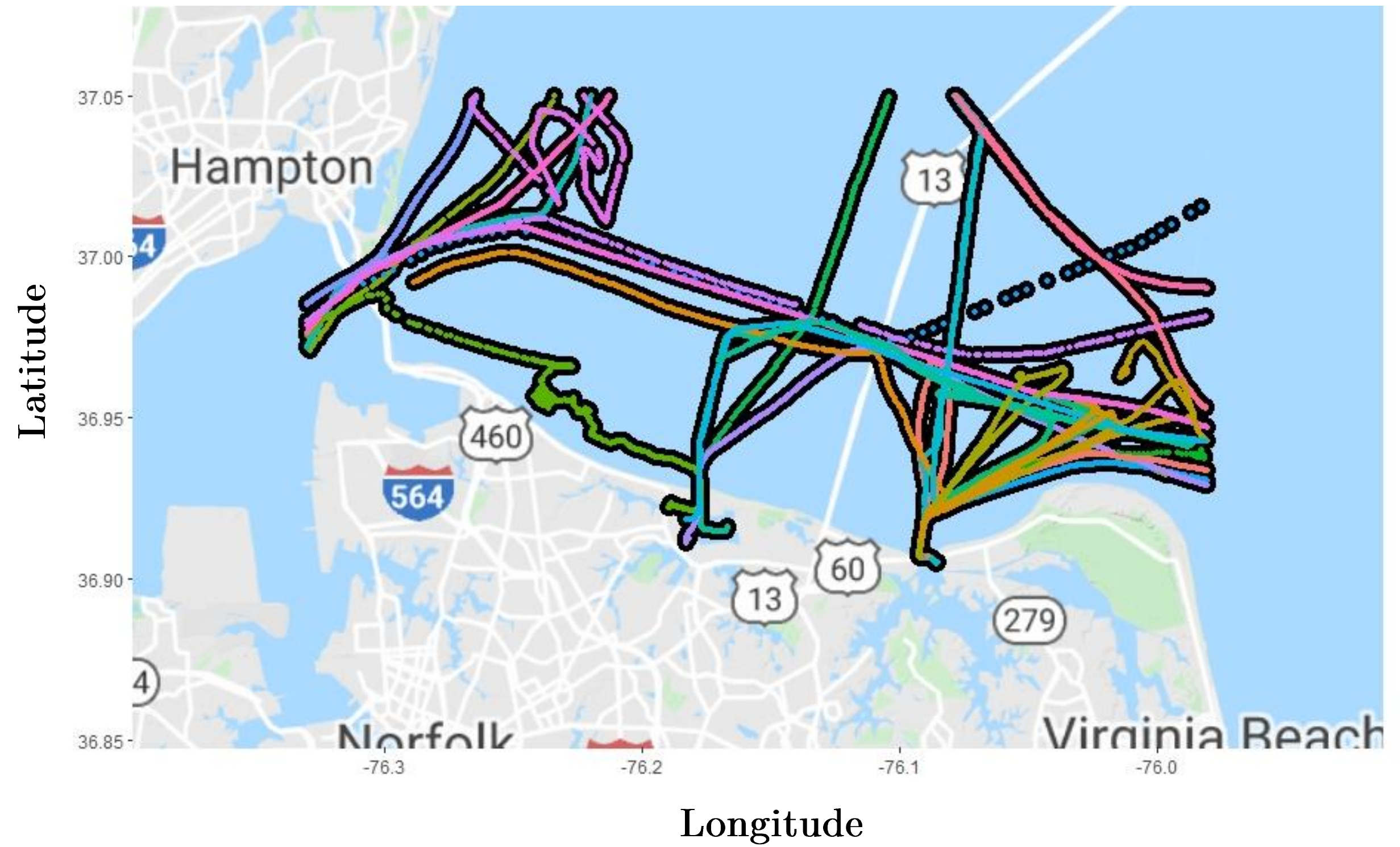}
        \label{fig:AISs5}
	\end{subfigure}
	\caption{Tracking performance of our algorithm. Thicker black dots represent the true tracks and thinner multicolored dots represent the associated tracks.}
	\label{fig:Totalv}
\end{figure*}

To visualize the track association outcome of our approach, we highlight the tracking performance of our algorithm for five test datasets in Fig.~\ref{fig:Totalv}. The true tracks are represented by thicker black dots whereas the associated tracks are represented by thinner multicolored dots.

Overall, we can say that our algorithm achieved superior performance across all metrics which are specifically designed to evaluate the performance of a track association approach in different aspects of association (i.e., identifying a new track, marking the end of a track, avoid mixing two different but close tracks that often cross each other, correctly identifying each segment of a track, etc.). We believe, it portrays the strength of our approach in dealing with complicated track association problems.


\subsection{Performance Comparison with MOTs} \label{sec:perf2}

We also implemented two popular MOT approaches, namely GNN (Global nearest neighbor) and JPDA (Joint Probabilistic Data Association) to evaluate their performances with respect to our association approach.  We summarized their performances in Table~\ref{tbl:10}-~\ref{tbl:12} to highlight their competitive edge. Among all the comparison cases, there is not a single instance that these MOT methods outperform our proposed method.  In many (albeit not all) cases, they are even worse than the sample algorithm (the baseline method) provided by the data challenge organizers.

\begin{table*}
\centering
\fontsize{9}{9}\selectfont
\caption{Counts of missed and broken tracks (separated by a ``;'') produced by the competing algorithms. Smaller the better. Bold indicates the best performance.}
\label{tbl:10}
\begin{tabular}{|l|l|l|l|l|l|}
\hline
Algorithms     & Test Set-1 & Test Set-2 & Test Set-3 & Test Set-4 & Test Set-5  \\ \hline
Our Algorithm  & \textbf{0};\textbf{0}      & \textbf{0};\textbf{0}      & \textbf{0};\textbf{0}      & \textbf{2};\textbf{2}     & \textbf{1};\textbf{1}      \\ \hline
Tracker GNN   & 3;12     &10;179    &7;470   & 11;284     &11;174       \\ \hline
Tracker JPDA  &3;708    &10;2233       & 3; 5775     & 8;5308      & 5; 6007     \\ \hline
Sample Algorithm  &1;0     &1;0       & 7;11     & 7;12     & 7;14      \\ \hline
\end{tabular}
\end{table*}

\begin{table*}
\centering
\fontsize{9}{9}\selectfont
\caption{Counts of extra, merged, and swapped tracks (separated by ``;'') produced by the competing algorithms. Smaller the better. Bold indicates the best performance.}
\label{tbl:11}
\begin{tabular}{|l|l|l|l|l|l|}
\hline
Algorithms     & Test Set-1 & Test Set-2 & Test Set-3 & Test Set-4 & Test Set-5  \\ \hline
Our Algorithm  &\textbf{0};\textbf{0};\textbf{0}      & \textbf{0};\textbf{0};\textbf{0}      & \textbf{1};\textbf{1};\textbf{169}      & \textbf{0};\textbf{0};\textbf{44}     & \textbf{1};\textbf{1};\textbf{1250}      \\ \hline
Tracker GNN   & 1;10;38     &17;179;1670    &4;467;3038    & 7;280;5532      &4;167;9887       \\ \hline
Tracker JPDA  &3;708;554    &2;2235;2437       & 5;5777;4025     &8;5308;11030      & 10;6012;13341     \\ \hline
Sample Algorithm  &6;5;729      &5;4;1296       &15;91;6484      &13;18;7953     & 15;22;8452      \\ \hline
\end{tabular}
\end{table*}

According to our observation, these MOTs lose their edge when the vessels are very close to one another, when tracks cross each other and when data collection frequency is high. They severely under-perform in the presence of time gaps and when the number of vessels to track is high (density of vessels in the surveillance area). These factors explain their poor association performance in test sets \#3, \#4 and \#5. In fact, all these characteristics made the challenge problems harder, and thus difficult for the existing algorithms to associate vessels to the true tracks correctly. The data challenge organizers, NGA, in collaboration with NSF, purposely included these tough characteristics in the challenge datasets in order to identify the most competent track association algorithms.

\begin{table*}
\centering
\fontsize{9}{9}\selectfont
\caption{Continuity and mean completeness scores (separated by a ``;'') produced by the competing algorithms. Values between 0 and 1. Greater the better.  Bold indicates the best performance.}
\label{tbl:12}
\begin{tabular}{|l|l|l|l|l|l|}
\hline
Algorithms    & Test Set-1 & Test Set-2 & Test Set-3 & Test Set-4 & Test Set-5  \\ \hline
Our Algorithm  &\textbf{1.00};\textbf{1.00}      & \textbf{1.00};\textbf{1.00}      & \textbf{0.95};\textbf{0.98}      & \textbf{0.99};\textbf{0.88}    & \textbf{1.00};\textbf{0.92}    \\ \hline
Tracker GNN   & 0.98;0.98    &0.88;0.77    &0.65;0.61    & 0.62;0.49      &0.78;0.49       \\ \hline
Tracker JPDA  &0.46;0.04    &0.39;0.01       &0.28;0.01     &0.10;0.01      & 0.19;0.01     \\ \hline
Sample Algorithm  &0.84;0.81      &0.94;0.85      &0.64;0.60      &0.76;0.63     & 0.70;0.61      \\ \hline
\end{tabular}
\end{table*}

\subsection{Performance Comparison with an Open Source AIS Dataset} \label{sec:perf34}

To further evaluate the efficacy of our proposed approach, we apply it to an additional open source AIS dataset~\cite{mult3:2022}. The dataset contains an extra variable named `heading', as compared to the AIS datasets used earlier. To be consistent, we discard this extra variable, so that our existing algorithm can be used without any further modification. To make the dataset size similar to what we have processed so far, we downsample this new dataset by considering first 25,000 instances. The downsampled dataset contains 186 vessels, much more than the number of vessels included in any datasets in the Data Association Challenge. We apply our algorithm to this downsampled dataset and compute the performance metrics. We summarize this performance evaluation in Table~\ref{tbl:100}, where we also listed the performance of the sample algorithm designed by the competition organizers. Our algorithm achieved the best performance across all metrics and outperformed the sample algorithm by a large margin. We want to mention particularly that our algorithm detected 182 vessels, 2\% different from the true number (186), whereas the sample algorithm detected 259 vessels, 40\% more than the true number.

\begin{table*}
\centering
\fontsize{9}{9}\selectfont
\caption{Performance comparison of our algorithm and the sample algorithm using the additional AIS dataset~\cite{mult3:2022}. Bold indicates the best performance.}
\label{tbl:100}
\begin{tabular}{|l|l|l|l|l|l|l|l|l|}
\hline
\multirow {3}{*}{Algorithms}   & Missed  & Extra & Merged & Broken & Swapped &  Continuity & Mean & Median \\
        &Tracks &Tracks & Tracks & Tracks & Tracks & Score &Completeness &Completeness \\
        &&&&&&&Score&Score\\ \hline
Our Algorithm    & \textbf{38}   & \textbf{34}  & \textbf{27}   & \textbf{23}   & \textbf{2,868}  & \textbf{0.93}    & \textbf{0.93}                   & \textbf{1}                         \\ \hline
Sample Algorithm & 104           & 177          & 59            & 132           & 15,552          & 0.35             & 0.33                            & 0.25   \\ \hline
\end{tabular}
\end{table*}

\section{CONCLUSION} \label{sec:conclusion}

In this paper, we propose a track association algorithm that associates the AIS data points to true tracks/clusters. Our research found that for the surface vessels, one can reasonably anticipate the next node location based on the AIS measurements taken at the last location of the vessel.  Contrasting the anticipated location and the current observed location is an effective strategy to carry out online track association. But a pure trajectory anticipation is not sufficient.  In order to have a robust track association, one has to deal with a number of operational complexities. For instance, as the geodesic distance between the anticipated and current locations gets greater, the effectiveness of the anticipation-based association deteriorates. To tackle these complexities, we spatially divide the area under consideration and devise adaptive policies that generate tracks depending on vessel locations. We also turned certain unique characteristics of maritime vessels and their movement into a set of guidelines that ensure a high success rate of the track association process. Using five test datasets provided by the NSF ATD data association challenge, we demonstrate that our algorithm produces superior tracking performance. Our approach also comes superior when compared with two state of art MOT approaches and evaluated on an open source AIS dataset.

The proposed approach is one of the very first attempts to associate trajectory observations to their tracks in real time using a spatio-temporal framework. We are confident that it will advance the trajectory analytics field beyond the regime of the traditional trajectory clustering approaches. The proposed approach can be used for the purpose of dynamic threat detection involving simultaneous movement of multiple objects. In the future study, more than using only the last node location in a number of decision steps, one can consider a small sequence of nodes, or the latent features extracted from them, to perform the track association process. Doing so could possibly help fix the residual problems associated with the accuracy of node location prediction.


\section*{Acknowledgment}

Imtiaz Ahmed and Mikyoung Jun acknowledge support by NSF DMS-1925119 under the NSF's Algorithms for Threat Detection (ATD) program.
Analysis results are derived from AIS data provided by the US Coast Guard at \url{https://www.navcen.uscg.gov}. The data was modified by the National Geospatial-Intelligence Agency in support of the National Science Foundation's ATD program.

\bibliographystyle{IEEEtranN}
\bibliography{ATD-2020AUG26bib}

\vspace{-1 cm}
\begin{IEEEbiography}[{\includegraphics[width=1in,height=1.25in,clip,keepaspectratio]{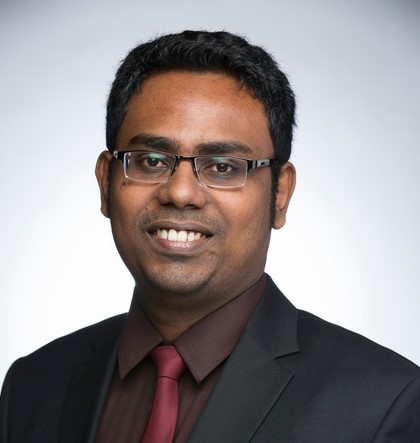}}]{Imtiaz Ahmed}
received B.Sc. and M.Sc. in Industrial \& Production Engineering from Bangladesh University of Engineering \& Technology, Dhaka, Bangladesh in 2012 and 2014 respectively. He received his Ph.D. in Industrial Engineering from Texas A\&M University in 2020. He is currently working as an Assistant Professor in the Department of Industrial \& Management Systems Engineering at West Virginia University. His research interests are in data analytics, machine learning and quality control.
\end{IEEEbiography}

\vspace{-1 cm}
\begin{IEEEbiography}[{\includegraphics[width=1in,height=1.25in,clip,keepaspectratio]{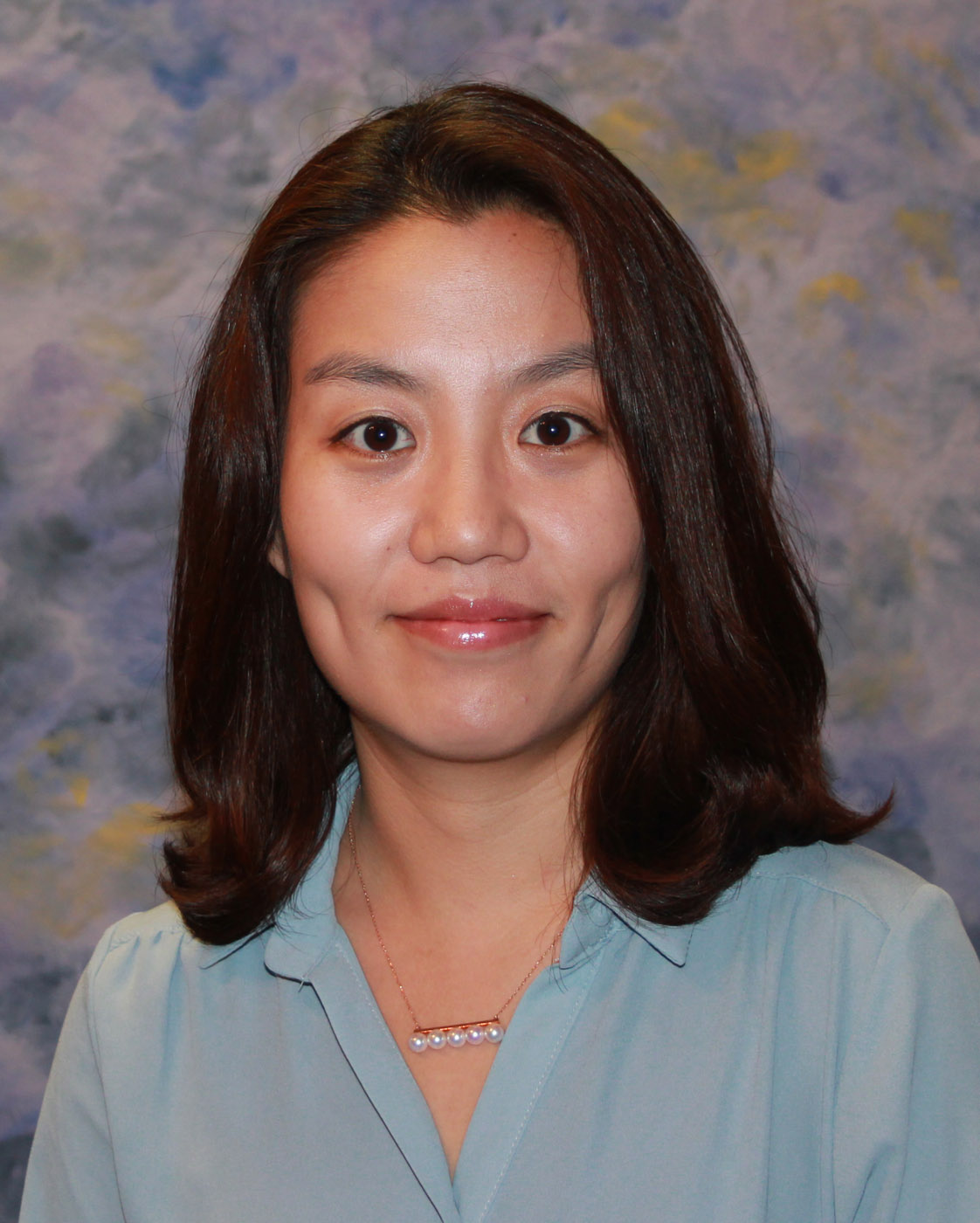}}]{Mikyoung Jun}received her Ph.D. in Statistics from University of Chicago (2005). She is currently a Professor and also a ConocoPhillips Data Science Professor of the Department of Mathematics at the University of Houston. One of her main research areas is spatio-temporal modeling for environmental and climate applications, especially development of parametric non-stationary and non-separable covariance functions for univariate, as well as multivariate processes. Dr. Jun is a fellow of ASA, and an elected member of ISI.
\end{IEEEbiography}

\vspace{-1 cm}
\begin{IEEEbiography}[{\includegraphics[width=1in,height=1.25in,clip,keepaspectratio]{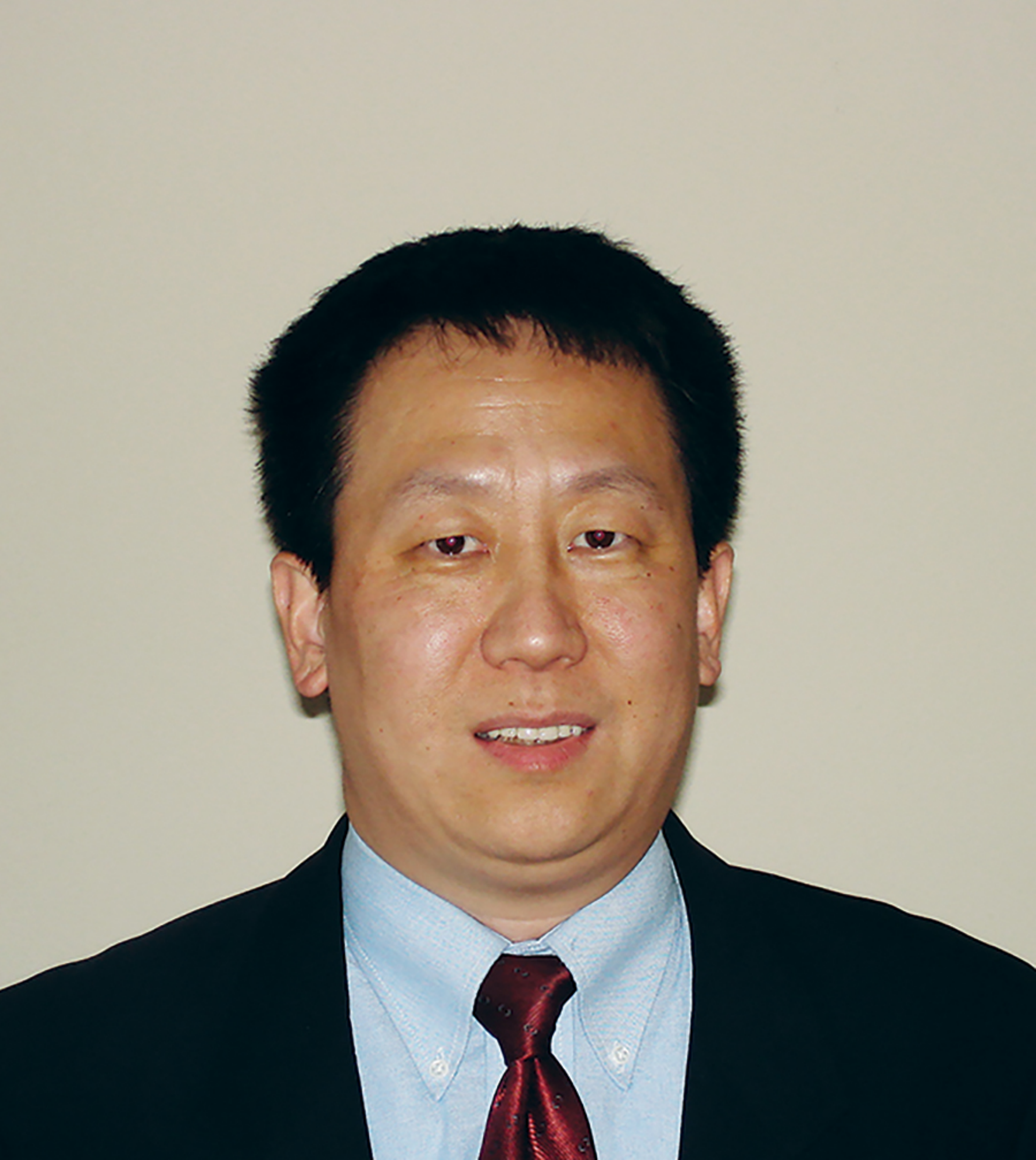}}]{Yu Ding}
(M'01, SM'11) received B.S. from University of Science \& Technology of China (1993); M.S. from Tsinghua University, China (1996); M.S. from Penn State University (1998); received Ph.D. in Mechanical Engineering from University of Michigan (2001). He is currently the Mike and Sugar Barnes Professor of Industrial \& Systems Engineering and a Professor of Electrical \& Computer Engineering at Texas A\&M University. His research interests are in quality and data science. Dr. Ding is a fellow of IIE, a fellow of ASME, a senior member of IEEE, and a member of INFORMS.
\end{IEEEbiography}


\end{document}